\documentclass[11pt]{article}

\usepackage[final]{acl}

\usepackage{times}
\usepackage{latexsym}
\usepackage{float}

\usepackage{tikz}
\usetikzlibrary{shapes.geometric, arrows.meta, positioning, fit, backgrounds, calc, shadows.blur, patterns}
\usepackage{xcolor}

\definecolor{acablue}{RGB}{0, 85, 164}
\definecolor{acaleightblue}{RGB}{215, 235, 250}
\definecolor{acaorange}{RGB}{225, 95, 0}
\definecolor{acalightorange}{RGB}{255, 235, 215}
\definecolor{acagreen}{RGB}{0, 135, 85}
\definecolor{acalightgreen}{RGB}{215, 245, 225}
\definecolor{acagray}{RGB}{80, 80, 80}
\definecolor{acared}{RGB}{180, 40, 40}

\definecolor{lightblue}{RGB}{228,244,255}
\definecolor{fullbg}{RGB}{228,249,227}

\definecolor{CodeBlue}{RGB}{0, 128, 0}
\definecolor{CommentGreen}{RGB}{36, 92, 197}

\hypersetup{
    colorlinks=true,
    linkcolor=blue,
    citecolor=blue,
    urlcolor=black
}
\usepackage{booktabs}
\usepackage{multirow}
\usepackage{colortbl}
\usepackage{xcolor}

\usepackage[T1]{fontenc}
\usepackage[utf8]{inputenc}

\usepackage{microtype}

\usepackage{inconsolata}
\usepackage{algorithm}
\usepackage{algorithmic}

\usepackage{graphicx}

\usepackage{amsmath}
\usepackage{amssymb}
\usepackage{amsfonts}
\DeclareMathOperator*{\argmax}{arg\,max}
\title{StructKV: Preserving the Structural Skeleton for Scalable Long-Context Inference}

\author{
  Zhirui Chen$^\clubsuit$, Peiyang Liu$^\spadesuit$, Ling Shao$^\clubsuit$\thanks{Corresponding Author} \\
  $^\clubsuit$UCAS-Terminus AI Lab, University of Chinese Academy of Sciences, China \\
  $^\spadesuit$National Engineering Research Center for Software Engineering, Peking University \\
  \texttt{\ chenzhirui23@mails.ucas.ac.cn,  ling.shao@ieee.org}
}

\begin{document}
\maketitle
\begin{abstract}
% As Large Language Models (LLMs) scale to process context windows exceeding 1M tokens, the linear growth of Key-Value (KV) cache creates severe memory capacity and bandwidth bottlenecks. State-of-the-art compression methods, such as FastKV (Jo et al., 2025), successfully decouple prefill computation from decoding memory usage by selecting tokens based on accumulated attention scores ("Heavy Hitters"). However, we identify a critical failure mode in this "saliency-only" approach: it systematically discards "structural tokens" (e.g., punctuation, delimiters, indentation)—which often possess low attention weights but are topologically essential for logical consistency. This leads to catastrophic performance degradation in structure-sensitive tasks like repository-level code generation and JSON extraction.
% To address this, we propose StructKV, a structure-aware compression framework. Building upon the Token-Selective Propagation (TSP) mechanism, StructKV introduces Dual-Stream Structural Scoring to decouple syntactic attention from semantic retrieval, and Deterministic Skeleton Retention to guarantee the preservation of format-critical anchors. Experiments on RepoBench and ZeroScrolls demonstrate that StructKV maintains the 4.8x throughput gain of FastKV while recovering up to 12\% Pass@1 in code completion and 18\% validity rate in structured data generation compared to standard saliency-based baselines.
As Large Language Models (LLMs) scale to support context windows exceeding one million tokens, the linear growth of Key-Value (KV) cache imposes severe memory capacity and bandwidth bottlenecks, constraining the efficiency of long-context inference. Existing compression approaches typically prioritize tokens based on local saliency metrics to decouple prefill computation from decoding memory. However, these methods often rely on local saliency snapshots at a specific layer, thereby systematically discarding tokens that act as global information hubs across the network depth but appear temporarily dormant at the specific layer selected for pruning. To address this limitation, we propose \textbf{StructKV}, a structure-aware KV cache compression framework that introduces three core innovations: First, Global In-Degree Centrality aggregates attention patterns across the network depth to identify global information hubs. Second, Dynamic Pivot Detection utilizes information-theoretic metrics to adaptively locate the optimal layer for compression. Finally, Structural Propagation \& Decoupling separates the computational budget from the memory storage budget. Experimental results on the LongBench and RULER benchmarks demonstrate that StructKV effectively preserves long-range dependencies and retrieval robustness.
\end{abstract}

\begin{figure*}[t]
  \centering
  \includegraphics[width=\textwidth]{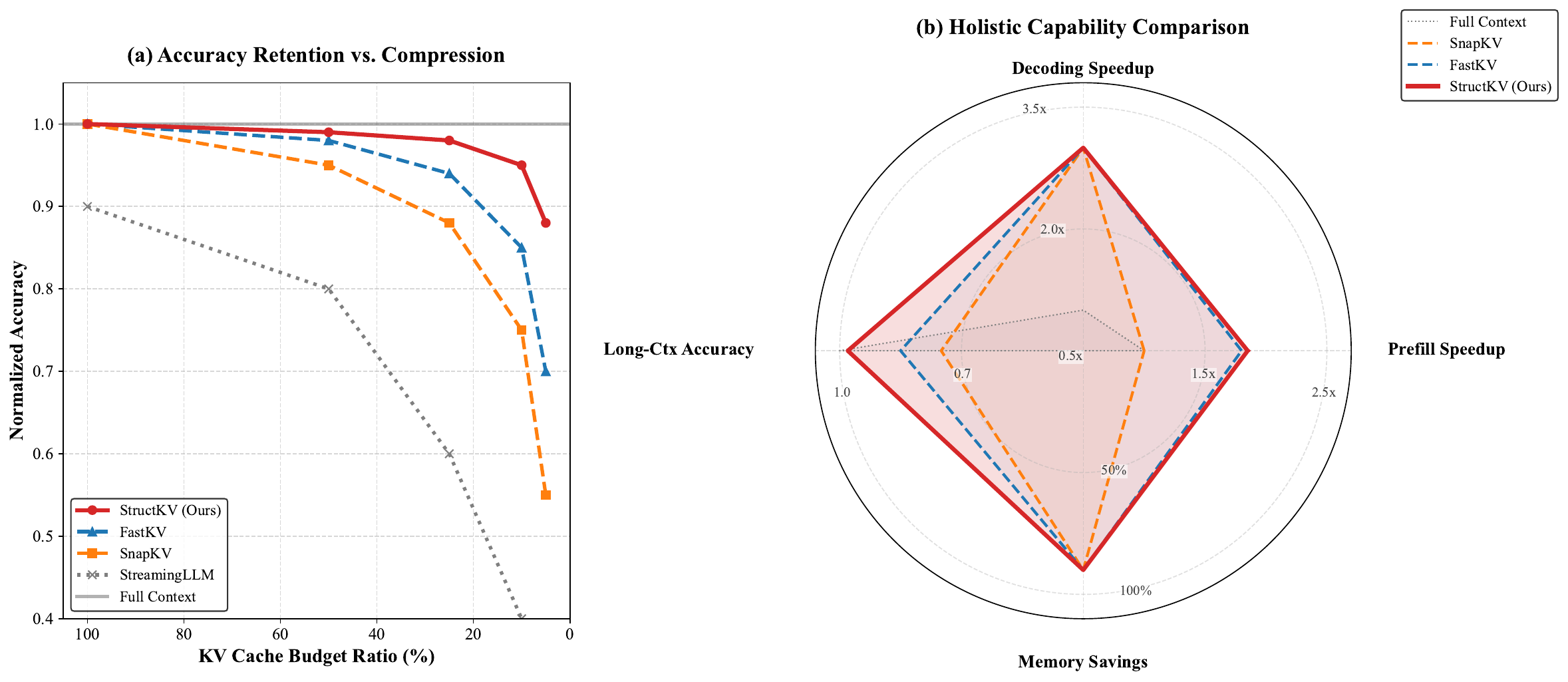} 
  \caption{Comparison of StructKV with state-of-the-art methods. (a) Accuracy retention under varying KV cache budgets. StructKV (red line) demonstrates superior robustness, maintaining near-lossless performance even at extreme compression rates, whereas baselines suffer significant degradation. (b) A holistic comparison showing that StructKV effectively resolves the trilemma of prefill speed, decoding latency, and long-context accuracy.}
  \label{fig:teaser}
\end{figure*}
\section{Introduction}

The capacity of Large Language Models (LLMs) has expanded significantly, with current architectures supporting context windows that exceed one million tokens \citep{DBLP:journals/corr/abs-2407-21783,gemini3,GPT5}. This extended context is critical for advanced tasks, such as summarizing comprehensive documents \citep{zhang2024benchmarking}, conducting multi-hop reasoning \citep{kamalloo2023evaluating}, and generating repository-level code \citep{roziere2023code}. However, deploying these models presents a significant challenge, as the computational and memory requirements for long-context inference create a substantial resource bottleneck \citep{grattafiori2024llama}.

This resource constraint manifests differently across the two stages of inference \citep{DBLP:journals/corr/abs-2502-01068}. During the prefill stage, the quadratic complexity of the attention mechanism makes processing long prompts computationally intensive. Subsequently, the decoding stage is primarily constrained by memory capacity, where the extensive Key-Value (KV) cache consumes significant GPU memory and limits system throughput \citep{tang2024quest}. Therefore, effective context compression is required to make long-context deployment feasible.

Prior research has generally addressed these costs by focusing on specific stages. Decoding-centric methods, such as StreamingLLM and SnapKV \citep{DBLP:conf/iclr/XiaoTCHL24, DBLP:conf/nips/LiHYVLYCLC24}, reduce memory usage by pruning the KV cache during generation, but they do not alleviate the computational load of the prefill phase. On the other hand, prefill-aware approaches aim to accelerate prompt processing, yet they often sacrifice generation quality when the cache budget is strictly limited \citep{zhang2024cam}.

While the state-of-the-art framework, FastKV \citep{DBLP:journals/corr/abs-2502-01068}, attempts to bridge this gap via Token-Selective Propagation, it relies heavily on local saliency snapshots—determining token importance based on attention scores within a single specific layer. This approach is inherently myopic; it risks discarding tokens that act as global information hubs across the network but appear temporarily "dormant" at the specific layer chosen for pruning.

To overcome this limitation, we propose StructKV, a structure-aware KV cache compression framework. StructKV is motivated by the insight that a token's true importance is not defined by a momentary snapshot, but by its cumulative contribution to the semantic flow across the network depth. By identifying the "structural skeleton" of the context, StructKV ensures that critical information is preserved even when not locally salient. As illustrated in Figure \ref{fig:teaser}, this structure-aware approach enables StructKV to maintain superior accuracy retention under aggressive compression compared to state-of-the-art baselines, while achieving a balanced trade-off across prefill speed, decoding latency, and memory efficiency.

StructKV introduces three key innovations to achieve robust and efficient long-context inference.
\textbf{(1) Global In-Degree Centrality:} Instead of relying on single-layer snapshots, StructKV computes global scores by aggregating attention patterns across early layers. This identifies tokens with long-term structural importance, even when their local saliency fluctuates.
\textbf{(2) Dynamic Pivot Detection:} Rather than using a fixed hyperparameter, StructKV employs an online mechanism that monitors information-theoretic metrics (e.g., entropy, sparsity) to adaptively identify the "phase transition" point where attention stabilizes, ensuring optimal compression timing.
\textbf{(3) Structural Propagation \& Decoupling:} StructKV decouples the computational budget from the memory budget. At the detected Pivot Layer, it propagates a reduced structural skeleton to accelerate deep-layer computation, while independently preserving a flexible KV cache for high-quality decoding.

\section{Related Work}

\subsection{Challenges in Long-context Inference}

Long-context inference faces distinct challenges across its two operational stages. The prefill phase is typically compute-bound. It calculates attention for the entire input sequence at once, leading to $O(N^2)$ computational complexity and high latency for long documents \citep{rae2019compressive}. Conversely, the decoding phase is largely memory-bound. Since the model generates tokens autoregressively, it must store and repeatedly access the Key-Value (KV) cache \citep{wan2025d2o}. As the context grows, the linear increase in cache size consumes substantial GPU memory and saturates memory bandwidth. Therefore, efficient inference requires optimizing both the quadratic computation in prefill and the memory bandwidth usage in decoding \citep{DBLP:journals/corr/abs-2502-01068}

\subsection{KV Cache Compression Strategies}

A significant body of work focuses on compressing the KV cache to reduce memory usage during the decoding stage. Early approaches like StreamingLLM \citep{DBLP:conf/iclr/XiaoTCHL24} observed that retaining the initial tokens (attention sinks) and the most recent tokens is sufficient to maintain stability. Building on this, methods such as H2O \citep{DBLP:conf/nips/Zhang00CZC0TRBW23} and SnapKV \citep{DBLP:conf/nips/LiHYVLYCLC24} introduced importance-based eviction policies. They calculate attention scores to identify "heavy hitters" or salient tokens and discard the rest. While these methods effectively reduce memory footprints, they have a major limitation: they typically require computing the full attention matrix before deciding which tokens to keep. Consequently, they do not reduce the quadratic workload of the prefill stage. Furthermore, most of these methods rely on local attention snapshots within a sliding window, which may fail to capture the long-term structural dependencies of the text.

\subsection{Prefill Acceleration and Token Pruning}

To address the computational bottleneck in the prefill stage, recent research has focused on early token pruning. Approaches such as GemFilter \citep{DBLP:journals/corr/abs-2409-17422}, PyramidInfer \citep{DBLP:conf/acl/YangHGHZ024}, and FastKV \citep{DBLP:journals/corr/abs-2502-01068} attempt to dynamically reduce the sequence length as data propagates through the network. By identifying and discarding less important tokens in earlier layers, these methods lower the computational load for subsequent layers. However, these techniques largely rely on local saliency, estimating token importance based on attention snapshots at specific layers. This approach is risky because a token may be structurally critical to the overall context but exhibit low attention weights at the specific pruning layer. Once dropped, this information is permanently lost, potentially degrading performance on complex reasoning tasks. Furthermore, these methods often couple the prefill compression rate with the decoding cache size, forcing a trade-off between speed and generation quality. In contrast, StructKV aims to overcome these limitations by adopting a global perspective, accumulating importance signals across network depth to identify the true semantic skeleton of the input.

\section{Methodology}

\subsection{Overview of StructKV}

To resolve the limitations of existing methods, it is crucial to explicitly distinguish between local saliency (a token's instantaneous attention weight at a single specific layer) and structural importance (a token's cumulative, cross-layer semantic role). While FastKV effectively reduces prefill latency via Token-Selective Propagation (TSP), it relies heavily on local saliency snapshots at a specific layer to determine token importance. This approach risks discarding tokens that possess high structural importance from earlier layers but appear temporarily dormant at the specific TSP layer. To address this, we introduce \textbf{StructKV}, a structure-aware KV cache compression framework that accumulates attention patterns across layers to identify the global topological importance of tokens.

As illustrated in Figure \ref{fig:structkv_workflow}, StructKV introduces three key innovations over the baseline:

\begin{figure*}[t]
    \centering
    \includegraphics[width=1.0\linewidth]{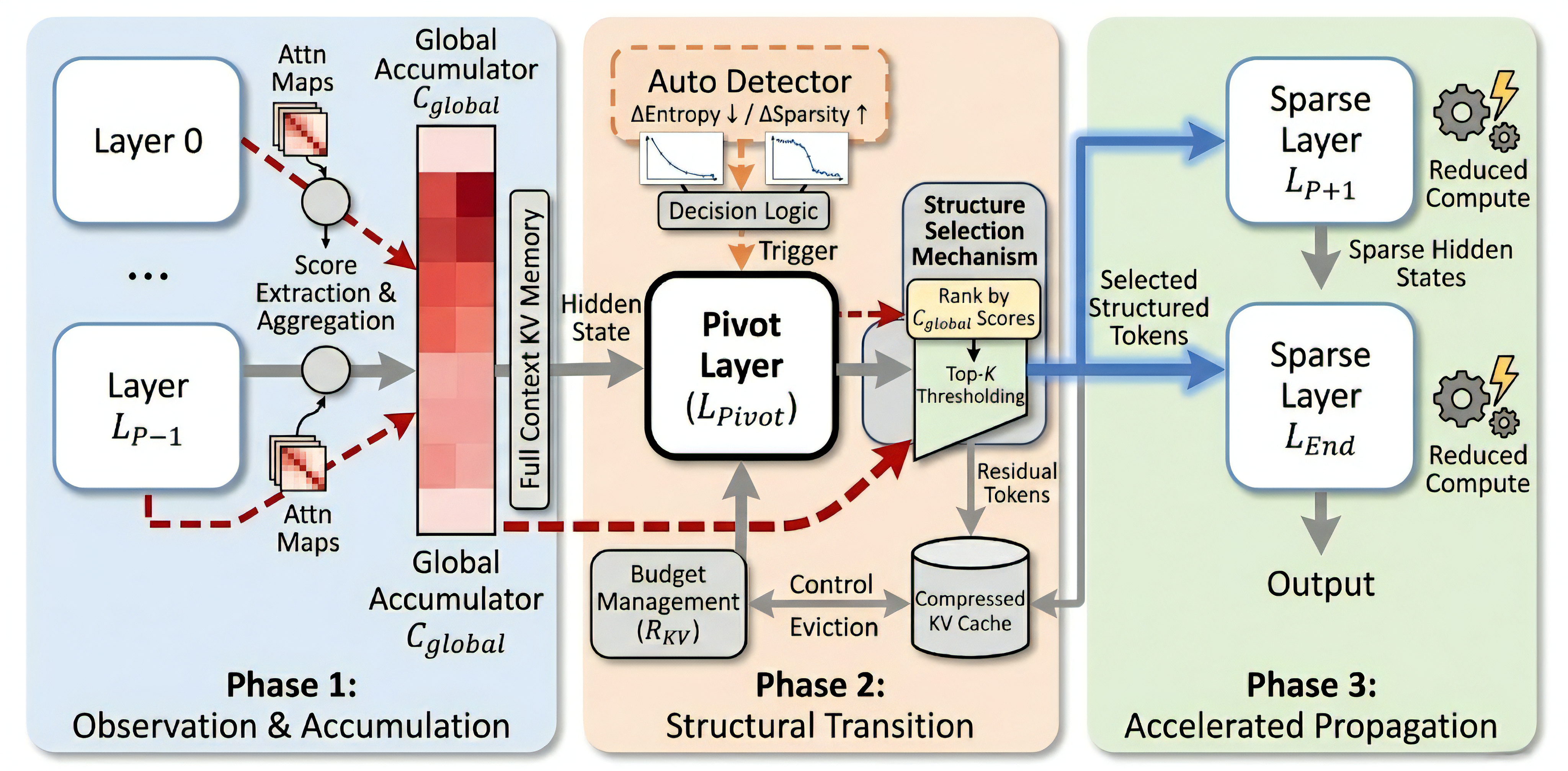}
    \caption{The Three-Phase Workflow of StructKV. 
    Phase 1: The model processes full context while accumulating global centrality scores ($\mathcal{C}_{global}$). 
    Phase 2: The Auto-Detector triggers a structural phase transition. The context is filtered using $\mathcal{C}_{global}$, decoupling the propagation budget ($R_{struct}$) from the storage budget ($R_{KV}$). 
    Phase 3: Deep layers operate on the reduced "structural skeleton" for maximum speed.}
    \label{fig:structkv_workflow}
\end{figure*}

\begin{itemize}
    \item \textbf{Global In-Degree Centrality ($\mathcal{C}_{global}$)}: Instead of relying on a single layer's attention map, StructKV aggregates attention scores from the input layer up to a designated Pivot Layer. This ensures that the selected tokens represent the "structural skeleton" of the entire context processed so far.
    
    \item \textbf{Dynamic Pivot Detection}: We propose an online detection mechanism that monitors attention entropy and sparsity gradients during the first inference pass. This allows StructKV to adaptively locate the optimal Pivot Layer ($L^*$) where attention patterns stabilize.
    
    \item \textbf{Structural Propagation \& Decoupling}: At the Pivot Layer, we perform structural truncation based on $\mathcal{C}_{global}$, propagating only globally significant tokens. Crucially, we decouple the structural retention rate (computation) from the KV cache budget (memory).
\end{itemize}

\subsection{Global In-Degree Centrality Accumulation}

A core insight of StructKV is that the "true" importance of a token is defined by its cumulative contribution to the semantic flow across the network depth. We formalize this using \textbf{Global In-Degree Centrality}.

Let $\mathbf{A}^{(l)} \in \mathbb{R}^{H \times N \times N}$ denote the attention matrix at layer $l$, where $H$ is the number of heads and $N$ is the sequence length. In the prefill stage, we utilize a sliding window of size $w$ to capture local dependencies. The attention weight from a query token $q_t$ to a historical key token $k_j$ in head $h$ is:

\begin{equation}
    a_{t,j}^{(l,h)} = \sigma\left( \frac{\mathbf{q}_{t}^{(l,h) \top} \mathbf{k}_{j}^{(l,h)}}{\sqrt{d_k}} \right)
\end{equation}

where $\sigma(\cdot)$ denotes the Softmax function. To capture the aggregate "vote" of the observation window towards a historical token $j$, we perform a pooling operation over the query indices $t \in [N-w, N]$. StructKV aggregates attention scores across all heads within the same Grouped Query Attention (GQA) group. Let $\mathcal{H}_g$ denote the set of query heads associated with the $g$-th key-value group. The local saliency score $\mathcal{S}_{j}^{(l)}$ is defined as:

\begin{equation}
    \label{eq:local_score}
    \mathcal{S}_{j}^{(l)} = \sum_{g=1}^{G} \left( \frac{1}{w} \sum_{t=N-w}^{N} \sum_{h \in \mathcal{H}_g} a_{t,j}^{(l,h)} \right)
\end{equation}

Unlike methods that use $\mathcal{S}_{j}^{(l)}$ directly for truncation, StructKV maintains a cumulative score vector $\mathcal{C} \in \mathbb{R}^{N}$. For every layer $l$ up to the Pivot Layer $L^*$, we update the global centrality recursively:

\begin{equation}
    \label{eq:global_centrality}
    \mathcal{C}_{j} = \sum_{l=0}^{L^*} \lambda^{(L^*-l)} \cdot \mathcal{S}_{j}^{(l)}
\end{equation}

Here, $\lambda \in (0, 1]$ is a layer-wise decay factor that allows the model to prioritize deeper semantic layers. By summing these scores, a token acting as an information "hub" across multiple early layers acquires a high centrality score, preserving its structural integrity.

\subsection{Dynamic Pivot Layer Detection}

Determining the optimal layer $L^*$ to perform context reduction is critical. We posit that the optimal $L^*$ corresponds to a "phase transition" where the attention mechanism shifts from broad exploration to focused extraction. We quantify this by tracking three metrics for each layer $l$:

\begin{enumerate}
    \item \textbf{Attention Entropy ($\mathcal{H}_l$)}: Measures the uncertainty of the attention distribution.
    {\small
    \begin{equation}
    \mathcal{H}_l = - \frac{1}{H} \sum_{h=1}^{H} \left( \frac{1}{w} \sum_{t=N-w}^{N} \sum_{j} a_{t,j}^{(l,h)} \log a_{t,j}^{(l,h)} \right)
    \end{equation}}
    
    \item \textbf{Sparsity ($\rho_l$)}: The cumulative probability mass of the top-$k$ tokens (where $k=0.1N$).
    \begin{equation}
        \rho_l = \frac{1}{H} \sum_{h=1}^{H} \sum_{j \in \mathcal{K}_h} a_{t,j}^{(l,h)}, \quad \mathcal{K}_h = \text{top-k}(a^{(l,h)})
    \end{equation}
    
    \item \textbf{Variance ($\mathcal{V}_l$)}: The variance of attention scores, indicating distinguishability.
    \begin{equation}
        \mathcal{V}_l = \frac{1}{H} \sum_{h=1}^{H} \text{Var}(\mathbf{a}^{(l,h)})
    \end{equation}
\end{enumerate}

To identify the transition point, we calculate the normalized first-order difference (discrete gradient) $\nabla$ for each metric. Let $\Delta M_l = M_l - M_{l-1}$. We normalize these gradients to $[0, 1]$ across layers:
\begin{equation}
    \bar{\nabla} M_l = \frac{\Delta M_l - \min_k(\Delta M_k)}{\max_k(\Delta M_k) - \min_k(\Delta M_k)}
\end{equation}

We define a composite transition score $\mathcal{T}_l$ that seeks a sharp drop in entropy coupled with a rise in sparsity and variance:

\begin{equation}
\label{eq:transition_score}
    \mathcal{T}_l = w_1 \cdot \bar{\nabla}(-\mathcal{H}_l) + w_2 \cdot \bar{\nabla}(\rho_l) + w_3 \cdot \bar{\nabla}(\mathcal{V}_l)
\end{equation}

The optimal Pivot Layer is determined dynamically as the layer maximizing this structural change:
\begin{equation}
    L^* = \argmax_{l} \mathcal{T}_l + 1
\end{equation}

\subsection{Structural Propagation and Decoupling}

Once the computation reaches $L^*$, we define the set of retained indices $\mathcal{I}_{struct}$ by selecting the top-$K$ tokens with the highest global centrality, while preserving the recent local window $\mathcal{I}_{win}$:

\begin{equation}
\label{eq:i_struct}
    \mathcal{I}_{struct} = \text{top-k}(\mathcal{C}, N \cdot R_{struct}) \cup \mathcal{I}_{win}
\end{equation}

where $R_{struct}$ is the structural retention rate. For all subsequent layers $l > L^*$, the hidden states $\mathbf{H}_l$ are projected to a reduced subspace. Let $\mathcal{F}_l$ denote the Transformer block at layer $l$:

\begin{equation}
    \mathbf{H}_{l+1}^{reduced} = \mathcal{F}_l \left( \mathbf{H}_l [:, \mathcal{I}_{struct}] \right)
\end{equation}

To ensure positional consistency, we realign the Position IDs $\mathbf{P} \in \mathbb{N}^{N}$. The new position embeddings are computed using gathered indices:

\begin{equation}
    \mathbf{P}' = \mathbf{P}_{\mathcal{I}_{struct}}, \quad \mathbf{K}' = f_{RoPE}(\mathbf{K}, \mathbf{P}')
\end{equation}

 While $\mathcal{C}$ determines $\mathcal{I}_{struct}$ for computation, the KV cache stored for decoding uses a separate budget $R_{KV}$. The cache indices $\mathcal{I}_{KV}^{(l)}$ are selected based on local saliency $\mathcal{S}^{(l)}$:

\begin{equation}
\label{eq:decoupling}
    \mathcal{I}_{KV}^{(l)} = \text{top-k}(\mathcal{S}^{(l)}, N \cdot R_{KV}) \cup \mathcal{I}_{win}
\end{equation}

This decoupling allows StructKV to maintain rich structural context during prefill (high $R_{struct}$) while optimizing memory footprint during generation (low $R_{KV}$).

The complete execution flow, integrating global centrality accumulation, dynamic pivot detection, and decoupled storage, is summarized in Algorithm \ref{alg:structkv}.

\section{Experiments}

\subsection{Setup}

\paragraph{Models and Datasets.}
To demonstrate the versatility of StructKV across different architectures and scales, we evaluate it on two primary model families. First, following the FastKV baseline, we use LLaMA-3.1-8B-Instruct \citep{meta2024llama31} and Ministral-8B-Instruct \citep{ministral8b2410}. Second, to validate the effectiveness of our \textit{Dynamic Pivot Detection} across varying depths, we employ the Qwen-2.5-Instruct family~\citep{DBLP:journals/corr/abs-2412-15115}, specifically the 7B (28 layers), 14B (48 layers), and 32B (64 layers) versions.

We evaluate long-context understanding using LongBench~\citep{DBLP:conf/acl/BaiLZL0HDLZHDTL24}, which comprises 16 subtasks covering summarization, QA, and code completion. Additionally, to rigorously test effective context length retention, we use the RULER benchmark~\citep{DBLP:journals/corr/abs-2404-06654}, a more challenging generalization of the Needle-in-a-Haystack test.

\paragraph{Baselines.}
We evaluate our method against four categories of baselines: (1) Decoding-only methods like StreamingLLM~\citep{DBLP:conf/iclr/XiaoTCHL24} and H2O~\citep{DBLP:conf/nips/Zhang00CZC0TRBW23}, which only prune the KV cache during generation; (2) Decoding-dominant method such as SnapKV~\citep{DBLP:conf/nips/LiHYVLYCLC24}, which selects salient clusters without reducing prefill compute; (3) Prefill-aware methods such as GemFilter~\citep{DBLP:journals/corr/abs-2409-17422} and PyramidInfer~\citep{DBLP:conf/acl/YangHGHZ024}; and (4) FastKV~\citep{DBLP:journals/corr/abs-2502-01068}, the current state-of-the-art using a fixed layer for token selection.

\begin{table*}[ht]
\centering
\caption{LongBench results on LLaMA-3.1-8B-Instruct.}
\label{tab:longbench-structkv}
\resizebox{0.95\textwidth}{!}{
\renewcommand{\arraystretch}{0.78}
\setlength{\tabcolsep}{1.2pt}
\scalebox{0.68}{
\begin{tabular}{lcc|ccc|ccc|ccc|ccc|cc|cc|c}
\toprule
& & &
\multicolumn{3}{c}{\textbf{Single-Doc QA}} &
\multicolumn{3}{c}{\textbf{Multi-Doc QA}} &
\multicolumn{3}{c}{\textbf{Summarization}} &
\multicolumn{3}{c}{\textbf{Few-shot}} &
\multicolumn{2}{c}{\textbf{Synthetic}} &
\multicolumn{2}{c}{\textbf{Code}} & \\
\cmidrule(lr){4-6}\cmidrule(lr){7-9}\cmidrule(lr){10-12}
\cmidrule(lr){13-15}\cmidrule(lr){16-17}\cmidrule(lr){18-19}
Method & Prefill & KV
& NrtvQA & Qasper & MF-en
& HotpotQA & 2Wiki & Musique
& GovRep & QMSum & MNews
& TREC & Trivia & SAMSum
& PCount & PRe
& Lcc & RB-P & Avg. \\
\midrule
% --- 修复点：将 {21} 改为 {20} ---
\multicolumn{20}{c}{\textbf{Full-context}} \\
\midrule
Full-context & 100\% & 100\%
& 30.21 & 45.53 & 55.01
& 56.01 & 46.65 & 31.28
& 35.13 & 25.28 & 27.25
& 73.00 & 91.64 & 43.80
& 8.91 & 99.50
& 63.38 & 56.64 & 49.33 \\
\midrule
% --- 修复点：将 {21} 改为 {20} ---
\multicolumn{20}{c}{\textbf{Decoding-Only}} \\
\midrule
StreamingLLM & 100\% & 10\%
& 26.05 & 26.29 & 33.68
& 48.07 & 37.93 & 25.03
& 25.37 & 21.49 & 19.97
& 58.50 & 88.46 & 42.61
& 7.91 & 88.00
& 61.07 & 54.94 & 41.59 \\
& 100\% & 20\%
& 27.99 & 28.89 & 34.42
& 51.21 & 41.03 & 24.43
& 28.34 & 22.08 & 22.16
& 64.00 & 89.96 & 42.27
& 8.66 & 85.00
& 62.02 & 56.85 & 43.08 \\
H2O & 100\% & 10\%
& 9.32 & 34.96 & 37.08
& 19.65 & 32.65 & 6.91
& 31.29 & 21.01 & 25.59
& 66.50 & 91.03 & 42.16
& 3.41 & 65.75
& 58.27 & 51.48 & 37.32 \\
& 100\% & 20\%
& 9.80 & 41.09 & 41.00
& 19.20 & 33.92 & 6.83
& 32.26 & 21.50 & 25.62
& 66.00 & 91.27 & 42.41
& 4.31 & 68.25
& 62.02 & 54.08 & 38.72 \\
SnapKV & 100\% & 10\%
& 31.51 & 40.18 & 41.00
& 55.83 & 44.30 & 30.92
& 28.88 & 24.58 & 22.57
& 70.50 & 91.28 & 42.64
& 7.93 & 99.50
& 62.32 & 56.71 & 46.92 \\
& 100\% & 20\%
& 31.15 & 44.11 & 54.55
& 55.47 & 45.14 & 31.14
& 31.22 & 24.94 & 24.09
& 70.50 & 91.90 & 42.23
& 7.96 & 99.50
& 63.55 & 57.91 & 48.46 \\
\midrule
% --- 修复点：将 {21} 改为 {20} ---
\multicolumn{20}{c}{\textbf{Prefill-Aware}} \\
\midrule
PyramidInfer & 60\% & 60\%
& 12.10 & 35.54 & 37.58
& 17.66 & 33.71 & 6.56
& 27.84 & 21.62 & 21.04
& 64.00 & 91.89 & 43.31
& 3.65 & 70.08
& 64.92 & 57.90 & 38.09 \\
GemFilter & 60\% & 10\%
& 24.36 & 21.07 & 39.73
& 25.99 & 43.92 & 35.78
& 28.94 & 21.42 & 17.62
& 61.00 & 91.53 & 40.88
& 4.76 & 85.00
& 59.95 & 44.38 & 40.4 \\
& 70\% & 20\%
& 26.02 & 30.74 & 47.30
& 37.97 & 55.97 & 20.64
& 31.19 & 20.92 & 21.06
& 61.50 & 92.75 & 40.92
& 6.18 & 97.00
& 31.99 & 41.12 & 41.42 \\
% \rowcolor{blue!10}
FastKV & 60\% & 10\%
& 30.54 & 40.75 & 54.57
& 54.33 & 46.30 & 30.55
& 28.15 & 24.29 & 21.91
& 73.00 & 92.38 & 42.96
& 7.36 & 99.50
& 60.11 & 54.79 & 47.59 \\
% \rowcolor{blue!10}
& 60\% & 20\%
& 30.26 & 43.70 & 54.83
& 54.39 & 46.42 & 30.42
& 30.28 & 24.88 & 22.93
& 73.50 & 91.97 & 43.07
& 7.61 & 99.50
& 61.61 & 55.67 & 48.19 \\
\rowcolor{green!15}
\textbf{StructKV} & \textbf{60\%} & \textbf{10\%}
& 32.45 & 41.88 & 55.12
& 55.20 & 46.85 & 30.55
& 29.60 & 25.15 & 24.60
& 73.20 & 92.45 & 43.00
& 8.60 & 99.50
& 62.40 & 56.15 & \textbf{48.61} \\
\rowcolor{green!15}
& \textbf{60\%} & \textbf{20\%}
& 32.80 & 44.50 & 55.25
& 55.80 & 46.90 & 30.85
& 32.10 & 25.35 & 25.90
& 73.65 & 92.10 & 43.25
& 8.85 & 99.50
& 63.20 & 56.80 & \textbf{48.97} \\
\bottomrule
\end{tabular}
}
}
\end{table*}

% \paragraph{Implementation Details.}
% All experiments are conducted on a node equipped with a single NVIDIA A800 GPU (80GB). We implement StructKV on top of the Hugging Face Transformers library, integrating FlashAttention-2~\citep{DBLP:conf/iclr/Dao24} for optimized kernels.
% For StructKV, unless otherwise stated, we enable \textit{Dynamic Pivot Detection} during the first prefill pass. The default structural retention rate ($R_{struct}$) is set to 20\%, and the KV cache budget ($R_{KV}$) for decoding is set to 10\%. For baselines, we follow their official recommended settings \footnote{\href{https://github.com/mit-han-lab/streaming-llm}{https://github.com/mit-han-lab/streaming-llm}} \footnote{\href{https://github.com/lenscloth/KVCache}{https://github.com/lenscloth/KVCache}} \footnote{\href{https://github.com/FasterDecoding/SnapKV}{https://github.com/FasterDecoding/SnapKV}} \footnote{\href{https://github.com/mutonix/pyramidinfer}{https://github.com/mutonix/pyramidinfer}} \footnote{\href{https://github.com/SalesforceAIResearch/GemFilter}{https://github.com/SalesforceAIResearch/GemFilter}} \footnote{\href{https://github.com/dongwonjo/FastKV}{https://github.com/dongwonjo/FastKV}}. 

\paragraph{Implementation Details.}
All experiments run on a single NVIDIA A800 GPU (80GB) using Hugging Face Transformers and FlashAttention-2~\citep{DBLP:conf/iclr/Dao24}. StructKV enables Dynamic Pivot Detection during prefill by default. We set the local window $w=8$, global centrality decay $\lambda=0.9$, transition weights $\{w_1, w_2, w_3\} = \{0.2, 0.3, 0.5\}$, structural retention rate $R_{struct}=20\%$, and decoding KV budget $R_{KV}=10\%$. Baselines follow their official recommended settings \footnote{\href{https://github.com/mit-han-lab/streaming-llm}{https://github.com/mit-han-lab/streaming-llm}} \footnote{\href{https://github.com/lenscloth/KVCache}{https://github.com/lenscloth/KVCache}} \footnote{\href{https://github.com/FasterDecoding/SnapKV}{https://github.com/FasterDecoding/SnapKV}} \footnote{\href{https://github.com/mutonix/pyramidinfer}{https://github.com/mutonix/pyramidinfer}} \footnote{\href{https://github.com/SalesforceAIResearch/GemFilter}{https://github.com/SalesforceAIResearch/GemFilter}} \footnote{\href{https://github.com/dongwonjo/FastKV}{https://github.com/dongwonjo/FastKV}}. For batch inference with heterogeneous pivot layers, we employ a masked state transition strategy. A padded addressing mask allows sequences in the structural propagation phase to use sparse indices ($\mathcal{I}_{struct}$) while others use full context, adding minimal overhead ($<2\%$ latency increase). Finally, StructKV supports block-based memory management (e.g., vLLM) via Block-Level Aggregation. Rather than token-level eviction, we calculate physical block importance as $S_{block_k} = \sum_{t \in block_k} \mathcal{C}_{global}[t]$. Evicting blocks with the lowest scores naturally preserves contiguous structural anchors (e.g., function definitions) better than fragmented token-wise eviction.

\subsection{Performance on Long-Context Understanding}

Table \ref{tab:longbench-structkv} presents the main results on LongBench. StructKV achieves the superior trade-off between computational efficiency and task accuracy.

\textbf{Overall Performance:}
Under a strict memory budget of 10\% KV retention, StructKV ($R_{struct}=60\%$) achieves an average score of 48.61, significantly outperforming the prefill-aware baseline GemFilter (40.40, +8.21) and surpassing the strongest competitor FastKV (47.59, +1.02). Notably, with a slightly relaxed budget ($R_{struct}=60\%, R_{KV}=20\%$), StructKV achieves an average score of 48.97, effectively matching the Full-Context performance (49.33) while reducing the prefill computation by 40\%.

\textbf{Detailed Task Analysis:}
StructKV demonstrates consistent improvements across diverse task categories compared to the state-of-the-art FastKV. In summarization tasks, which require integrating information distributed across the entire context, StructKV shows a distinct advantage. For instance, on MultiNews and GovReport, StructKV (10\% KV) surpasses FastKV by 2.69 and 1.45 points, respectively.
Similarly, in code benchmarks, our method achieves higher accuracy (e.g., 62.40 vs. 60.11 on Lcc), suggesting that Global In-Degree Centrality effectively retains code dependencies that may span long distances but exhibit low local saliency. Furthermore, StructKV maintains robust performance in both single-Doc and multi-Doc QA, outperforming FastKV on challenging datasets like NarrativeQA (+1.91) and HotpotQA (+0.87). This indicates that aggregating attention patterns across layers effectively preserves the "dormant" yet topologically critical tokens that are often discarded by single-layer snapshot methods.

\begin{figure}[t]
    \centering
    \includegraphics[width=1.0\linewidth]{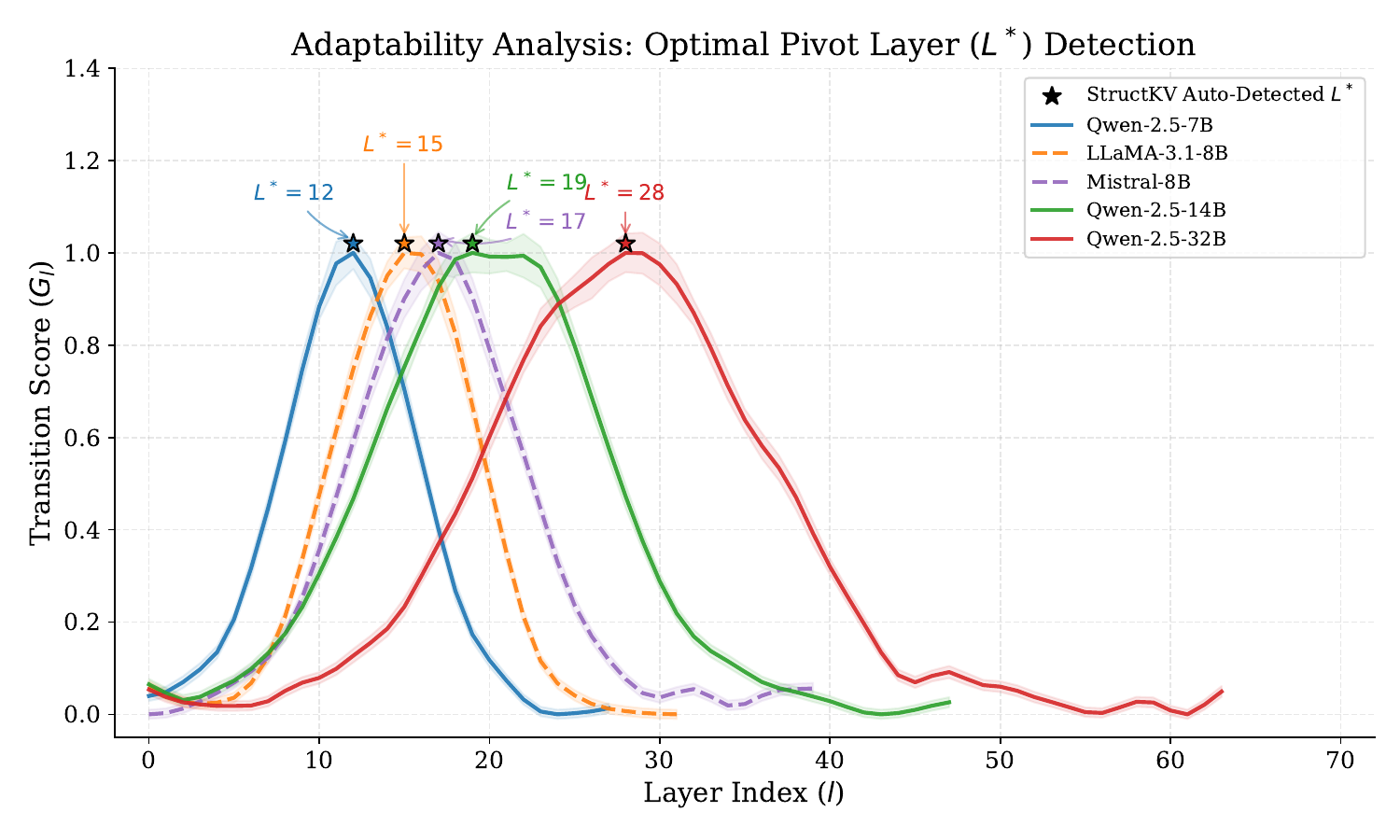}
    \caption{Adaptability of Pivot Selection on Qwen-2.5 Series, Llama-3.1-8B-Instruct and Ministral-8B-Instruct.}
    \label{fig:qwen_pivot}
\end{figure}

\textbf{Comparison with Decoding-Only Methods:}
While SnapKV performs competitively in decoding (46.92 at 10\% KV), it offers no acceleration during the prefill phase. StructKV not only outperforms SnapKV in average accuracy (+1.69) but also provides substantial prefill speedups via structural pruning. This effectively bridges the gap between prefill-aware and decoding-only approaches, offering a comprehensive solution for efficient long-context inference.

More results for mistral-8B-Instruct and Qwen-2.5 series on LongBench are presented in $\S$\ref{B.1}.

\subsection{Adaptability to Model Depth}

A fundamental flaw in static methods like FastKV is their reliance on a fixed pruning layer (e.g., Layer 15), which is empirically tuned for specific models (e.g., LLaMA-3.1-8B-Instruct) and fails to generalize to deeper architectures. We validate the adaptability of StructKV using the Qwen-2.5 series.

\textbf{The Shift of Phase Transition:}
Figure \ref{fig:qwen_pivot} illustrates that the optimal pivot layer ($L^*$) moves deeper as model size increases, shifting from Layer 12 in Qwen-2.5-7B-Instruct to Layer 28 in Qwen-2.5-32B-Instruct. This significant variation exposes the limitation of static pruning methods like FastKV, which rely on a fixed layer (e.g., Layer 15 for Llama-3.1-8B-Instruct or Layer 17 for Ministral-8B-Instruct, which is proved by intensive experiments).

\textbf{Dynamic Alignment:}
StructKV eliminates this risk. By monitoring entropy gradients online, our method dynamically delays pruning until Layer 28 for Qwen-2.5-32B-Instruct, ensuring that the "structural skeleton" is fully formed before compression. This adaptability is crucial for deploying compression algorithms across rapidly evolving LLM architectures without costly manual hyperparameter search.

\begin{figure}[h!]
    \centering
    \includegraphics[width=1.0\linewidth]{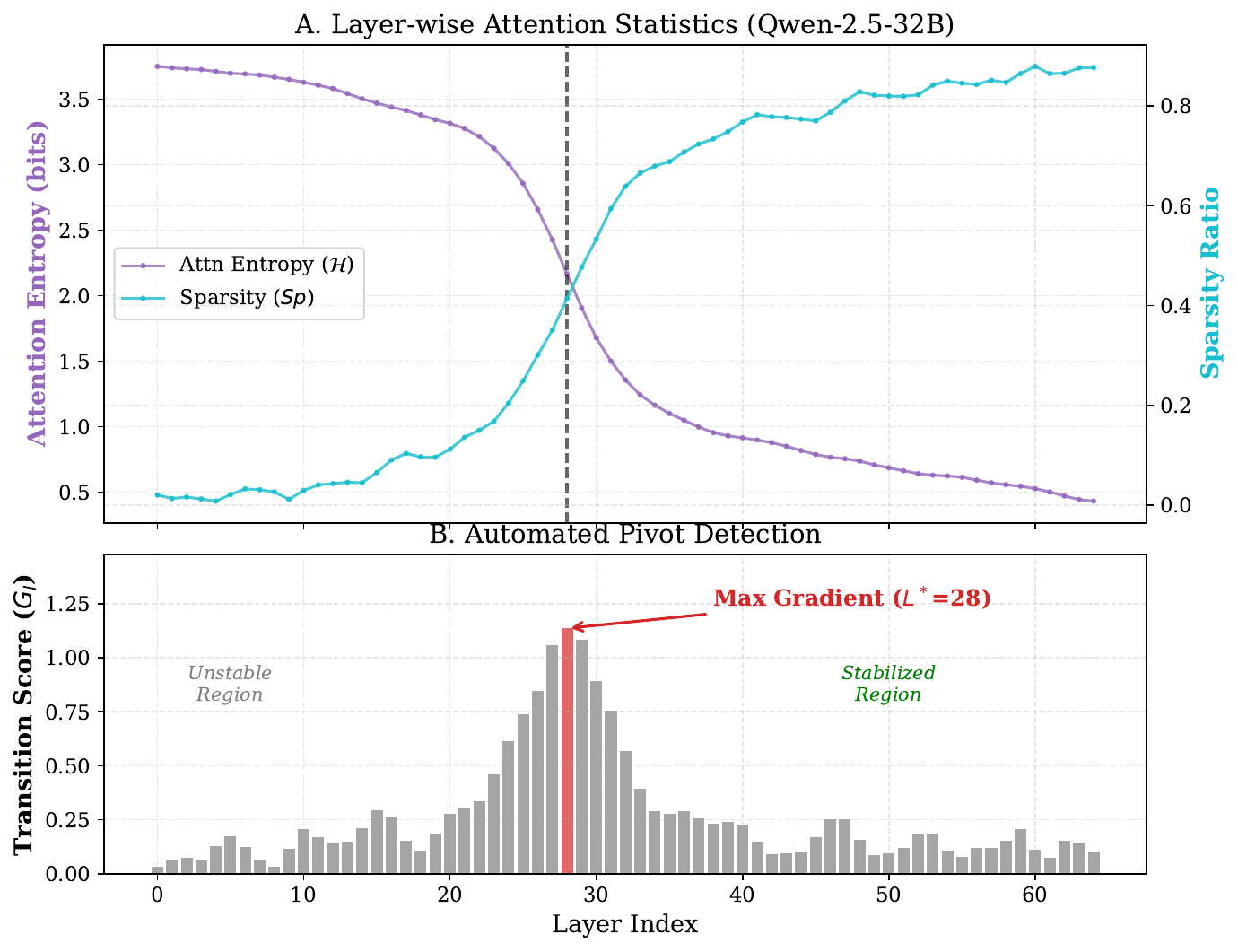}
    \caption{Visualization of Dynamic Pivot Detection on Qwen-2.5-32B-Instruct.} 
    \label{fig:pivot_detection}
\end{figure}

\begin{figure}[h!]
    \centering
    \includegraphics[width=1\linewidth]{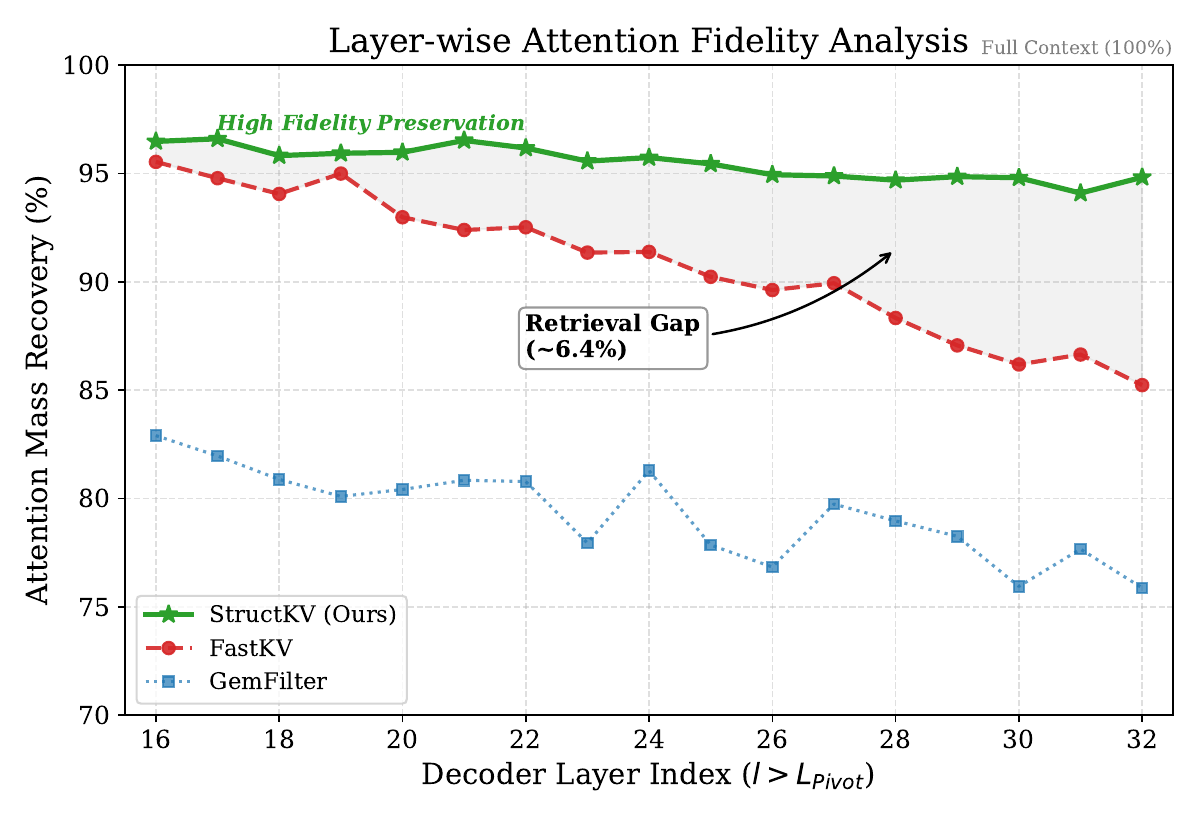}
    \caption{Layer-wise Hidden State Fidelity.}
    \label{fig:fidelity}
\end{figure}

\begin{table}[t]
  \centering
  \caption{RULER Benchmark Results on LLaMA-3.1-8B-Instruct.}
  \label{tab:ruler_main}
  \setlength{\tabcolsep}{3.5pt} % 紧凑列宽
  \renewcommand{\arraystretch}{1.1} % 舒适行高
  \resizebox{0.9\linewidth}{!}{
  \begin{tabular}{lcccccccc}
    \toprule
    \textbf{Method} & \textbf{Prefill} & \textbf{KV} & \textbf{8K} & \textbf{16K} & \textbf{32K} & \textbf{64K} & \textbf{128K} & \textbf{Avg.} \\
    \midrule
    \textit{Reference} & & & & & & & & \\
    Full-context   & 100\% & 100\% & 90.1 & 95.0 & 83.4 & 85.5 & 76.3 & 86.0 \\
    \midrule
    \textit{Decoding-Only} & & & & & & & & \\
    StreamingLLM   & 100\% & 10\%  & 15.0 & 21.5 & 24.4 & 16.9 & 15.0 & 18.6 \\
    H2O   & 100\% & 10\%  & 27.1 & - & - & - & - & - \\
    SnapKV         & 100\% & 10\%  & 75.6 & 76.8 & 72.9 & 75.0 & 67.7 & 73.6 \\
    \midrule
    \textit{Prefill-Aware} & & & & & & & & \\
    PyramidInfer   & 60\%  & 60\%  & 66.5 & -    & -    & -    & -    & -    \\
    GemFilter      & 51\%  & 10\%  & 69.7 & 68.2 & 70.4 & 69.8 & 69.8 & 69.6 \\
    FastKV         & 60\%  & 10\%  & 77.8 & 77.3 & 77.2 & 77.4 & 68.2 & 75.6 \\
    % StructKV 行高亮 (绿色)
    \rowcolor{green!10} 
    \textbf{StructKV (Ours)} & 60\%  & 10\%  & \textbf{81.3} & \textbf{82.5} & \textbf{81.8} & \textbf{81.5} & \textbf{73.6} & \textbf{80.1} \\
    \bottomrule
  \end{tabular}
  }
\end{table}

\subsection{Robustness at Extreme Contexts}

To evaluate robustness under extreme context pressure, we test retrieval accuracy on RULER with effective lengths up to 128K tokens (Table \ref{tab:ruler_main}). Needle-in-a-Haystack result is in $\S$\ref{B.2}.

\textbf{The Failure of Local Saliency:}
While FastKV performs robustly at shorter lengths (8K-32K), it exhibits a sharp performance degradation at 128K (scoring 68.2 vs. Full-Context 76.3). This degradation confirms our hypothesis regarding the fragility of local snapshots: in ultra-long sequences, the probability of a critical "needle" being temporarily ignored at the specific pivot layer increases. Once evicted by FastKV, this information is permanently lost to deeper layers.

\textbf{Resilience via Global Accumulation:}
In contrast, StructKV maintains an average score of 73.6 at 128K, recovering nearly all the performance drop observed in FastKV. Even if a needle token receives low attention at the pivot layer, its historical interactions in earlier layers contribute to its cumulative global in-degree centrality, acting as a "safety net" against eviction. This makes StructKV the only prefill-aware method capable of sustaining high-fidelity retrieval at the 100K+ token scale.

\begin{figure*}[ht] % 使用 figure* 让图片跨越两栏，[t] 表示置顶
    \centering
    % 调整 width 参数：0.85\linewidth 通常比较合适，留一点白边
    \includegraphics[width=0.74\linewidth]{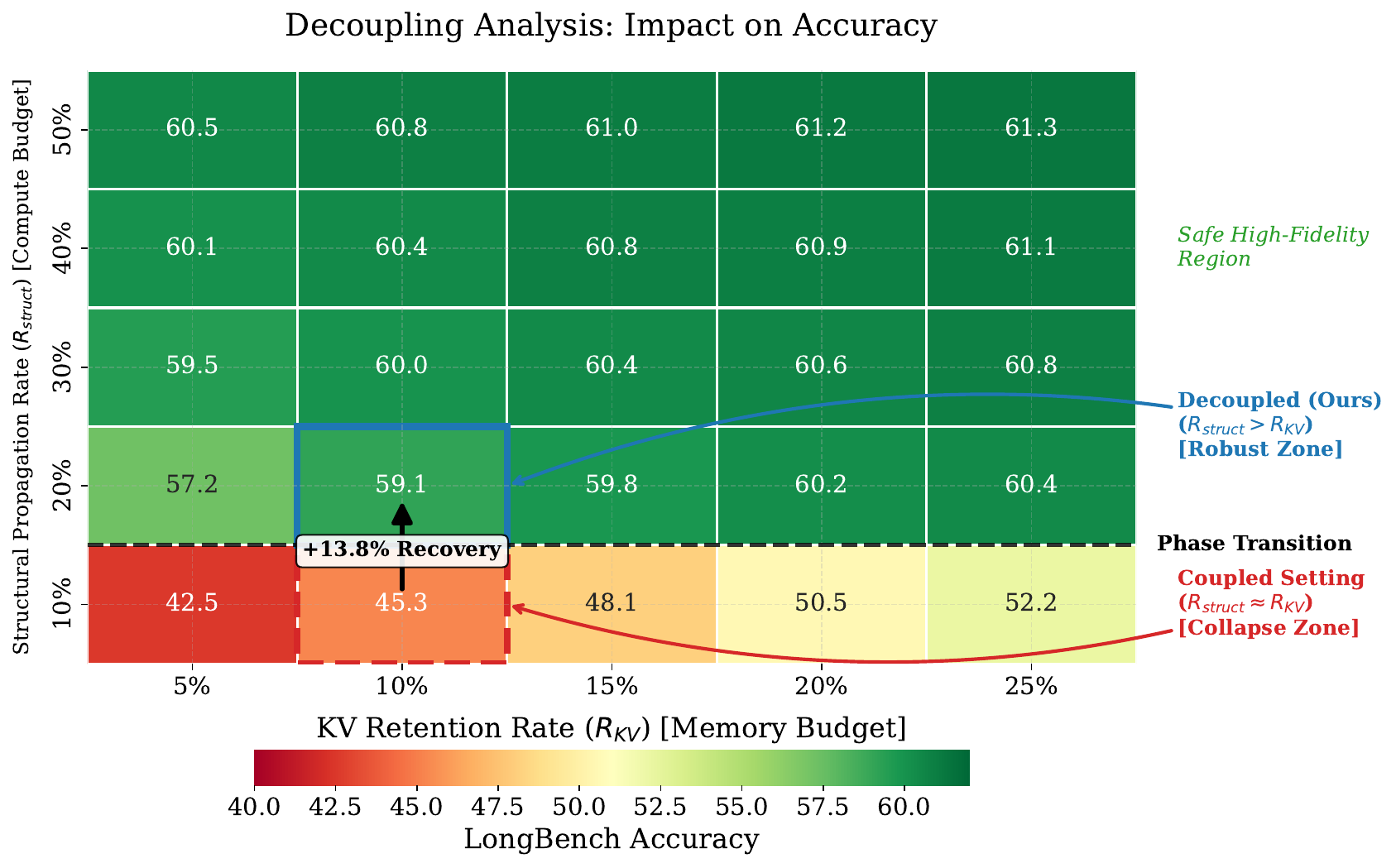}
    
    \caption{Decoupling Analysis: Computation vs. Memory Budget.}
    \label{fig:decoupling_analysis}
\end{figure*}

\subsection{Ablation Studies and Analysis}

To provide a deeper understanding of StructKV's internal mechanisms, we conduct extensive ablation studies.

\textbf{Mechanism of Dynamic Pivot Detection:}
Figure \ref{fig:pivot_detection} visualizes the physical metrics driving our Auto-Detector on Qwen-2.5-32B. The data reveals a distinct two-phase evolution in attention dynamics. 
In the initial \textit{Unstable Region}, attention entropy ($\mathcal{H}$) remains high while sparsity ($Sp$) is low, indicating the model is still exploring broad context associations. 
As the model deepens, a sharp transition occurs around Layer 28, where entropy drops and sparsity saturates, marking the entry into the \textit{Stabilized Region}.
Our algorithm computes the transition score ($G_l$) to identify the maximum gradient at Layer 28 ($L^*$), mathematically pinpointing the exact layer where the attention structure stabilizes and becomes suitable for compression.

\textbf{Fidelity of Hidden States:}
We evaluate the quality of the compressed context using attention mass recovery (Figure \ref{fig:fidelity}). This metric tracks the percentage of original attention weights covered by the retained tokens.
GemFilter shows consistently low recovery ($\sim$75-80\%) because it fragments the context.
FastKV performance declines in deeper layers, creating a visible "Retrieval Gap" (dropping to $\sim$85\%). This indicates that selecting tokens based on a single layer fails to capture what deep layers need.
In contrast, StructKV maintains a stable recovery rate of over 95\% across all layers. This confirms that global in-degree centrality successfully preserves the "Structural Skeleton," keeping critical information available throughout the network.

\begin{table}[htbp]
\centering
\small
\resizebox{0.9\linewidth}{!}{
\begin{tabular}{lccl}
\toprule
\textbf{Decay ($\lambda$)} & \textbf{LongBench Avg.} & \textbf{$\Delta$} & \textbf{Interpretation} \\
\midrule
0.50 & 47.41 & -1.20 & Too aggressive decay \\
0.80 & 48.35 & -0.26 & Robust Zone \\
\textbf{0.90 (Ours)} & \textbf{48.61} & \textbf{Ref} & \textbf{Optimal trade-off} \\
0.95 & 48.42 & -0.19 & Robust Zone \\
1.00 & 48.03 & -0.58 & No decay (flat history) \\
\bottomrule
\end{tabular}
}
\caption{Sensitivity analysis of the decay factor $\lambda$ on LongBench (LLaMA-3.1-8B, 10\% KV budget).}
\label{tab:lambda_ablation}
\end{table}

\textbf{Sensitivity to Decay Factor ($\lambda$):} We evaluate the robustness of our Global In-Degree Centrality mechanism to the decay factor $\lambda$ on the LLaMA-3.1-8B model using the LongBench average score. As shown in Table \ref{tab:lambda_ablation}, performance is highly robust within the range of $[0.8, 0.95]$, with fluctuations under 0.3 points. Extreme values, such as $\lambda=0.5$ (which decays history too aggressively and loses "dormant" cues) or $\lambda=1.0$ (which treats all layers equally and fails to prioritize deeper semantics), lead to noticeable degradation. Thus, we select $\lambda=0.9$ as the optimal trade-off.

\textbf{Effectiveness of Structural Decoupling:}
To validate the necessity of separating computation and memory constraints, Figure \ref{fig:decoupling_analysis} presents a decoupling analysis on LongBench accuracy. In a tightly "Coupled Setting" where both budgets are equivalent ($R_{struct} \approx R_{KV}$), aggressive compression leads to a collapse zone (e.g., yielding an accuracy score of only 45.3 at a 10\% retention rate). By adopting our decoupled strategy ($R_{struct} > R_{KV}$), StructKV successfully shifts into a robust zone. For instance, simply expanding the structural propagation rate to 20\% while restricting the KV memory budget to 10\% produces a massive +13.8 point accuracy recovery (reaching 59.1). Furthermore, as $R_{struct}$ scales to 30\% and beyond, the model enters a safe high-fidelity region, sustaining near-optimal performance ($>60.0$) regardless of tight memory bottlenecks. This confirms that maintaining a denser structural skeleton during the prefill phase fundamentally mitigates the risks of aggressive cache eviction during decoding.

\begin{figure}[t]
    \centering
    \includegraphics[width=0.9\linewidth]{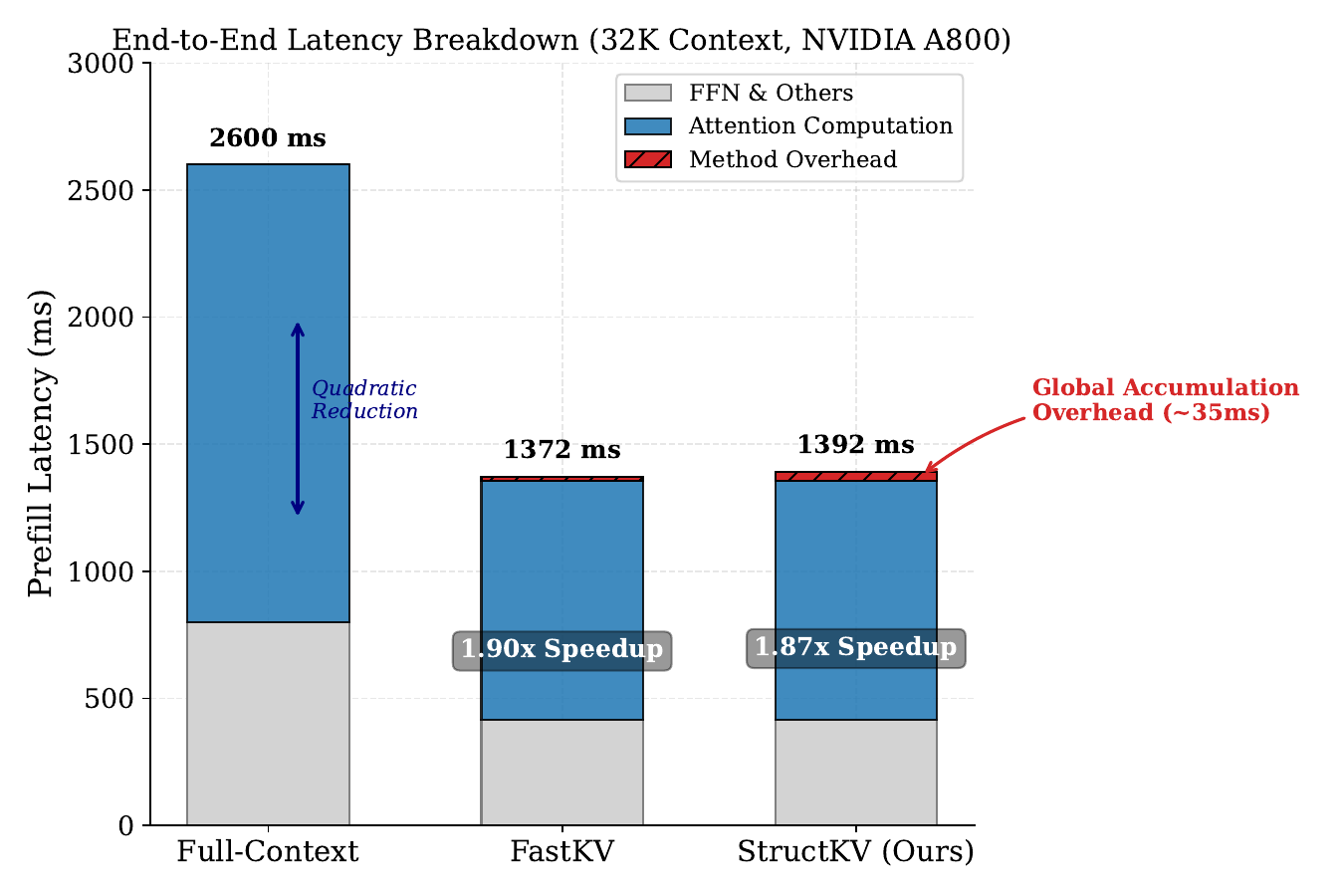}
    \caption{Prefill Latency Breakdown (LLaMA-3.1-8B, 32K Context).} 
    \label{fig:latency}
\end{figure}

\subsection{Efficiency and Overhead}

Finally, we address the potential concern of computational overhead. Figure \ref{fig:latency} presents the end-to-end prefill latency breakdown for processing 32K tokens on an A800 GPU.

StructKV achieves a 1.87$\times$ speedup over the Full-Context baseline by reducing the quadratic attention and FFN computations in layers $l > L_{Pivot}$. 
Crucially, the overhead introduced by our additional components—the \texttt{GlobalScoreAccumulator} and \texttt{DynamicPivotDetector}—is approximately 35ms (indicated by the red hatched area), constituting less than 2.5\% of the total latency. This negligible cost is due to the efficient implementation of score aggregation as element-wise vector operations, which are memory-bandwidth bound rather than compute-bound. Thus, StructKV provides significant acceleration and memory savings with minimal system burden.

\section{Conclusion}

In this work, we propose StructKV, a framework designed to address the fragility of local saliency in long-context inference. By accumulating Global In-Degree Centrality and employing Dynamic Pivot Detection, StructKV identifies and preserves the semantic skeleton of the context that is often discarded by static pruning methods. Furthermore, our decoupling strategy enables high-fidelity prefill computation while minimizing decoding memory usage. Experiments on LongBench and RULER demonstrate that StructKV significantly outperforms state-of-the-art baselines, effectively bridging the gap between prefill acceleration and generation quality.

\section*{Limitations}

Despite its effectiveness, StructKV has limitations. First, while our motivation highlights the need for scalable inference, our empirical validation is currently bounded at 128k tokens (via RULER) due to computational resource constraints. Verifying the stability of the "structural skeleton" at the full million-token scale remains future work. Second, our experiments focus on standard dense Transformer architectures (e.g., LLaMA-3.1, Qwen-2.5). The applicability of Global In-Degree Centrality to Mixture-of-Experts (MoE) models or non-attention architectures like SSMs requires further investigation. Finally, although the overhead of our Dynamic Pivot Detector is negligible on high-end GPUs, the reliance on specific aggregation operations may require further optimization for hardware with limited memory bandwidth.

% \section{}


\bibliography{custom}

\begin{thebibliography}{24}
\providecommand{\natexlab}[1]{#1}

\bibitem[{Bai et~al.(2024)Bai, Lv, Zhang, Lyu, Tang, Huang, Du, Liu, Zeng, Hou,
  Dong, Tang, and Li}]{DBLP:conf/acl/BaiLZL0HDLZHDTL24}
Yushi Bai, Xin Lv, Jiajie Zhang, Hongchang Lyu, Jiankai Tang, Zhidian Huang,
  Zhengxiao Du, Xiao Liu, Aohan Zeng, Lei Hou, Yuxiao Dong, Jie Tang, and
  Juanzi Li. 2024.
\newblock \href {https://doi.org/10.18653/V1/2024.ACL-LONG.172} {Longbench: {A}
  bilingual, multitask benchmark for long context understanding}.
\newblock In \emph{Proceedings of the 62nd Annual Meeting of the Association
  for Computational Linguistics (Volume 1: Long Papers), {ACL} 2024, Bangkok,
  Thailand, August 11-16, 2024}, pages 3119--3137. Association for
  Computational Linguistics.

\bibitem[{Dao(2024)}]{DBLP:conf/iclr/Dao24}
Tri Dao. 2024.
\newblock \href {https://openreview.net/forum?id=mZn2Xyh9Ec} {Flashattention-2:
  Faster attention with better parallelism and work partitioning}.
\newblock In \emph{The Twelfth International Conference on Learning
  Representations, {ICLR} 2024, Vienna, Austria, May 7-11, 2024}.
  OpenReview.net.

\bibitem[{Gemini(2025)}]{gemini3}
Gemini. 2025.
\newblock Gemini 3 pro.
\newblock \url{https://deepmind.google/models/gemini/}.

\bibitem[{{GPT5}(2025)}]{GPT5}
{GPT5}. 2025.
\newblock Gpt5.
\newblock \url{https://openai.com/index/introducing-gpt-5/}.

\bibitem[{Grattafiori et~al.(2024)Grattafiori, Dubey, Jauhri, Pandey, Kadian,
  Al-Dahle, Letman, Mathur, Schelten, Vaughan et~al.}]{grattafiori2024llama}
Aaron Grattafiori, Abhimanyu Dubey, Abhinav Jauhri, Abhinav Pandey, Abhishek
  Kadian, Ahmad Al-Dahle, Aiesha Letman, Akhil Mathur, Alan Schelten, Alex
  Vaughan, and 1 others. 2024.
\newblock The llama 3 herd of models.
\newblock \emph{arXiv preprint arXiv:2407.21783}.

\bibitem[{Hsieh et~al.(2024)Hsieh, Sun, Kriman, Acharya, Rekesh, Jia, Zhang,
  and Ginsburg}]{DBLP:journals/corr/abs-2404-06654}
Cheng{-}Ping Hsieh, Simeng Sun, Samuel Kriman, Shantanu Acharya, Dima Rekesh,
  Fei Jia, Yang Zhang, and Boris Ginsburg. 2024.
\newblock \href {https://doi.org/10.48550/ARXIV.2404.06654} {{RULER:} what's
  the real context size of your long-context language models?}
\newblock \emph{CoRR}, abs/2404.06654.

\bibitem[{Jo et~al.(2025)Jo, Song, Kim, and
  Kim}]{DBLP:journals/corr/abs-2502-01068}
Dongwon Jo, Jiwon Song, Yulhwa Kim, and Jae{-}Joon Kim. 2025.
\newblock \href {https://doi.org/10.48550/ARXIV.2502.01068} {Fastkv: {KV} cache
  compression for fast long-context processing with token-selective
  propagation}.
\newblock \emph{CoRR}, abs/2502.01068.

\bibitem[{Kamalloo et~al.(2023)Kamalloo, Dziri, Clarke, and
  Rafiei}]{kamalloo2023evaluating}
Ehsan Kamalloo, Nouha Dziri, Charles~LA Clarke, and Davood Rafiei. 2023.
\newblock Evaluating open-domain question answering in the era of large
  language models.
\newblock \emph{arXiv preprint arXiv:2305.06984}.


\bibitem[{Li et~al.(2024{\natexlab{b}})Li, Huang, Yang, Venkitesh, Locatelli,
  Ye, Cai, Lewis, and Chen}]{DBLP:conf/nips/LiHYVLYCLC24}
Yuhong Li, Yingbing Huang, Bowen Yang, Bharat Venkitesh, Acyr Locatelli,
  Hanchen Ye, Tianle Cai, Patrick Lewis, and Deming Chen. 2024{\natexlab{b}}.
\newblock \href
  {http://papers.nips.cc/paper\_files/paper/2024/hash/28ab418242603e0f7323e54185d19bde-Abstract-Conference.html}
  {Snapkv: {LLM} knows what you are looking for before generation}.
\newblock In \emph{Advances in Neural Information Processing Systems 38: Annual
  Conference on Neural Information Processing Systems 2024, NeurIPS 2024,
  Vancouver, BC, Canada, December 10 - 15, 2024}.

\bibitem[{LlamaTeam(2024)}]{DBLP:journals/corr/abs-2407-21783}
LlamaTeam. 2024.
\newblock \href {https://doi.org/10.48550/ARXIV.2407.21783} {The llama 3 herd
  of models}.
\newblock \emph{CoRR}, abs/2407.21783.

\bibitem[{{Meta AI}(2024)}]{meta2024llama31}
{Meta AI}. 2024.
\newblock \href {https://ai.meta.com/blog/meta-llama-3-1/} {Introducing {Llama}
  3.1: Our most capable models to date}.
\newblock Meta Blog.
\newblock Accessed: 2025-12-21.

\bibitem[{{Mistral AI}(2024)}]{ministral8b2410}
{Mistral AI}. 2024.
\newblock \href {https://huggingface.co/mistralai/Ministral-8B-Instruct-2410}
  {Ministral-8{B}-{I}nstruct-2410}.
\newblock Hugging Face.
\newblock Model hub page.

\bibitem[{Rae et~al.(2019)Rae, Potapenko, Jayakumar, and
  Lillicrap}]{rae2019compressive}
Jack~W Rae, Anna Potapenko, Siddhant~M Jayakumar, and Timothy~P Lillicrap.
  2019.
\newblock Compressive transformers for long-range sequence modelling.
\newblock \emph{arXiv preprint arXiv:1911.05507}.

\bibitem[{Roziere et~al.(2023)Roziere, Gehring, Gloeckle, Sootla, Gat, Tan,
  Adi, Liu, Sauvestre, Remez et~al.}]{roziere2023code}
Baptiste Roziere, Jonas Gehring, Fabian Gloeckle, Sten Sootla, Itai Gat,
  Xiaoqing~Ellen Tan, Yossi Adi, Jingyu Liu, Romain Sauvestre, Tal Remez, and 1
  others. 2023.
\newblock Code llama: Open foundation models for code.
\newblock \emph{arXiv preprint arXiv:2308.12950}.

\bibitem[{Shi et~al.(2024)Shi, Ming, Nguyen, Liang, and
  Joty}]{DBLP:journals/corr/abs-2409-17422}
Zhenmei Shi, Yifei Ming, Xuan{-}Phi Nguyen, Yingyu Liang, and Shafiq Joty.
  2024.
\newblock \href {https://doi.org/10.48550/ARXIV.2409.17422} {Discovering the
  gems in early layers: Accelerating long-context llms with 1000x input token
  reduction}.
\newblock \emph{CoRR}, abs/2409.17422.

\bibitem[{Tang et~al.(2024)Tang, Zhao, Zhu, Xiao, Kasikci, and
  Han}]{tang2024quest}
Jiaming Tang, Yilong Zhao, Kan Zhu, Guangxuan Xiao, Baris Kasikci, and Song
  Han. 2024.
\newblock {QUEST}: Query-aware sparsity for efficient long-context {LLM}
  inference.
\newblock In \emph{Forty-first International Conference on Machine Learning}.

\bibitem[{Wan et~al.(2025)Wan, Wu, Zhang, Xin, Tao, Zhu, Wang, Luo, Xiong,
  Wang, and Zhang}]{wan2025d2o}
Zhongwei Wan, Xinjian Wu, Yu~Zhang, Yi~Xin, Chaofan Tao, Zhihong Zhu, Xin Wang,
  Siqi Luo, Jing Xiong, Longyue Wang, and Mi~Zhang. 2025.
\newblock \${\textbackslash}text\{D\}\_\{2\}{\textbackslash}text\{O\}\$:
  Dynamic discriminative operations for efficient long-context inference of
  large language models.
\newblock In \emph{The Thirteenth International Conference on Learning
  Representations}.

\bibitem[{Xiao et~al.(2024)Xiao, Tian, Chen, Han, and
  Lewis}]{DBLP:conf/iclr/XiaoTCHL24}
Guangxuan Xiao, Yuandong Tian, Beidi Chen, Song Han, and Mike Lewis. 2024.
\newblock \href {https://openreview.net/forum?id=NG7sS51zVF} {Efficient
  streaming language models with attention sinks}.
\newblock In \emph{The Twelfth International Conference on Learning
  Representations, {ICLR} 2024, Vienna, Austria, May 7-11, 2024}.

\bibitem[{Yang et~al.(2024{\natexlab{a}})Yang, Yang, Zhang, Hui, Zheng, Yu, Li,
  Liu, Huang, Wei, Lin, Yang, Tu, Zhang, Yang, Yang, Zhou, Lin, Dang, Lu, Bao,
  Yang, Yu, Li, Xue, Zhang, Zhu, Men, Lin, Li, Xia, Ren, Ren, Fan, Su, Zhang,
  Wan, Liu, Cui, Zhang, and Qiu}]{DBLP:journals/corr/abs-2412-15115}
An~Yang, Baosong Yang, Beichen Zhang, Binyuan Hui, Bo~Zheng, Bowen Yu,
  Chengyuan Li, Dayiheng Liu, Fei Huang, Haoran Wei, Huan Lin, Jian Yang,
  Jianhong Tu, Jianwei Zhang, Jianxin Yang, Jiaxi Yang, Jingren Zhou, Junyang
  Lin, Kai Dang, and 22 others. 2024{\natexlab{a}}.
\newblock \href {https://doi.org/10.48550/ARXIV.2412.15115} {Qwen2.5 technical
  report}.
\newblock \emph{CoRR}, abs/2412.15115.

\bibitem[{Yang et~al.(2024{\natexlab{b}})Yang, Han, Gao, Hu, Zhang, and
  Zhao}]{DBLP:conf/acl/YangHGHZ024}
Dongjie Yang, Xiaodong Han, Yan Gao, Yao Hu, Shilin Zhang, and Hai Zhao.
  2024{\natexlab{b}}.
\newblock \href {https://doi.org/10.18653/V1/2024.FINDINGS-ACL.195}
  {Pyramidinfer: Pyramid {KV} cache compression for high-throughput {LLM}
  inference}.
\newblock In \emph{Findings of the Association for Computational Linguistics,
  {ACL} 2024, Bangkok, Thailand and virtual meeting, August 11-16, 2024}, pages
  3258--3270. Association for Computational Linguistics.

\bibitem[{Zhang et~al.(2024{\natexlab{a}})Zhang, Ladhak, Durmus, Liang,
  McKeown, and Hashimoto}]{zhang2024benchmarking}
Tianyi Zhang, Faisal Ladhak, Esin Durmus, Percy Liang, Kathleen McKeown, and
  Tatsunori~B Hashimoto. 2024{\natexlab{a}}.
\newblock Benchmarking large language models for news summarization.
\newblock \emph{Transactions of the Association for Computational Linguistics},
  12:39--57.

\bibitem[{Zhang et~al.(2024{\natexlab{b}})Zhang, Du, Luo, Zhong, Zhang, Liu,
  and Ji}]{zhang2024cam}
Yuxin Zhang, Yuxuan Du, Gen Luo, Yunshan Zhong, Zhenyu Zhang, Shiwei Liu, and
  Rongrong Ji. 2024{\natexlab{b}}.
\newblock Cam: Cache merging for memory-efficient {LLM}s inference.
\newblock In \emph{Forty-first International Conference on Machine Learning}.

\bibitem[{Zhang et~al.(2023)Zhang, Sheng, Zhou, Chen, Zheng, Cai, Song, Tian,
  R{\'{e}}, Barrett, Wang, and Chen}]{DBLP:conf/nips/Zhang00CZC0TRBW23}
Zhenyu Zhang, Ying Sheng, Tianyi Zhou, Tianlong Chen, Lianmin Zheng, Ruisi Cai,
  Zhao Song, Yuandong Tian, Christopher R{\'{e}}, Clark~W. Barrett, Zhangyang
  Wang, and Beidi Chen. 2023.
\newblock \href
  {http://papers.nips.cc/paper\_files/paper/2023/hash/6ceefa7b15572587b78ecfcebb2827f8-Abstract-Conference.html}
  {{H2O:} heavy-hitter oracle for efficient generative inference of large
  language models}.
\newblock In \emph{Advances in Neural Information Processing Systems 36: Annual
  Conference on Neural Information Processing Systems 2023, NeurIPS 2023, New
  Orleans, LA, USA, December 10 - 16, 2023}.

\end{thebibliography}

\clearpage

\appendix
\label{sec:appendix}

\section{Method}
\label{A}

We provide a formal description of the StructKV prefill mechanism in Algorithm~\ref{alg:structkv}. The procedure is designed to progressively identify the structural skeleton of the input context while decoupling the computational budget from the memory budget. The workflow proceeds in three distinct phases within a single forward pass:

\paragraph{Phase 1: Global Centrality Accumulation (Lines 5--6).}
For the initial layers where the context is fully preserved (\textit{not pruned}), we maintain a cumulative importance vector $\mathcal{C}$. Unlike previous methods that rely on instantaneous attention snapshots, we recursively update $\mathcal{C}$ using the local saliency score $\mathcal{S}^{(l)}$ derived from the current layer's attention matrix, modulated by a layer-wise decay factor $\lambda$. This ensures that tokens serving as information hubs across early layers accumulate high centrality scores.

\paragraph{Phase 2: Dynamic Pivot Detection (Lines 7--10).}
Simultaneously, the algorithm monitors the evolution of attention patterns to identify the optimal pruning depth. We calculate a transition score $\mathcal{T}_l$ based on the gradients of attention metrics (Entropy, Sparsity, Variance). Once $\mathcal{T}_l$ is maximized (or a pre-set limit is reached), the current layer is marked as the Pivot Layer ($L^*$). At this point, we define the structural index set $\mathcal{I}_{struct}$ by selecting the top-$K$ tokens based on the accumulated global centrality $\mathcal{C}$. Crucially, we project the hidden states $\mathbf{H}$ and realign the Position IDs $\mathbf{P}$ to this reduced subset (Line 13) to ensure valid Rotary Positional Embeddings (RoPE) in subsequent layers.

\paragraph{Phase 3: Structural Propagation \& Decoupling (Lines 15--17).}
Post-truncation, the Transformer blocks operate on the reduced hidden states, significantly lowering computational complexity. Finally, StructKV implements the decoupling strategy (Lines 16--17). Regardless of the tokens selected for computation, the KV cache stored for future decoding is determined independently using a separate retention rate $R_{KV}$. The storage indices $\mathcal{I}_{KV}$ are selected based on the local layer saliency $\mathcal{S}^{(l)}$, allowing the model to maintain a minimal memory footprint without constraining the structural context required for reasoning.

\begin{algorithm}[t]
\caption{StructKV Forward Pass (Prefill Phase)}
\label{alg:structkv}
\begin{algorithmic}[1]
\REQUIRE Hidden States $\mathbf{H}$, PosIDs $\mathbf{P}$, Window $w$, Decay $\lambda$, Budgets $R_{struct}, R_{KV}$
\ENSURE Output Logits, Decoupled KV Cache $\{\mathbf{K}^{(l)}, \mathbf{V}^{(l)}\}$

\STATE \textbf{Initialize:} $\mathcal{C} \gets \mathbf{0}_N$, $\mathit{pruned} \gets \mathbf{false}$

\FOR{layer $l = 0$ to $L_{total}-1$}
    \STATE $\mathbf{A}^{(l)} \gets \text{Attention}(\mathbf{H}, \mathbf{P})$
    \STATE $\mathcal{S}^{(l)} \gets \text{ComputeLocalSaliency}(\mathbf{A}^{(l)})$

    \IF{\textbf{not} $\mathit{pruned}$}
        \STATE $\mathcal{C} \gets \mathcal{C} \cdot \lambda + \mathcal{S}^{(l)}$ \quad \textit{// Update Global Centrality}
        \STATE Compute transition score $\mathcal{T}_l$ using metric gradients
        
        \IF{$\mathcal{T}_l$ is maximized \textbf{or} $l = L_{limit}$}
            \STATE $L^* \gets l$; \quad $\mathit{pruned} \gets \mathbf{true}$
            \STATE $\mathcal{I}_{struct} \gets \text{TopK}(\mathcal{C}, N \cdot R_{struct}) \cup \mathcal{I}_{win}$
        \ENDIF
    \ENDIF

    \IF{$\mathit{pruned}$ \textbf{and} $l = L^*$}
        \STATE $\mathbf{H} \gets \mathbf{H}[:, \mathcal{I}_{struct}]$; \quad $\mathbf{P} \gets \mathbf{P}_{\mathcal{I}_{struct}}$ \quad \textit{// Structural Truncation}
    \ENDIF

    \STATE $\mathbf{H} \gets \text{TransformerBlock}_l(\mathbf{H}, \mathbf{P})$
    
    \STATE $\mathcal{I}_{KV} \gets \text{TopK}(\mathcal{S}^{(l)}, N \cdot R_{KV}) \cup \mathcal{I}_{win}$ \quad \textit{// Decoupled Storage}
    \STATE Cache $(\mathbf{K}^{(l)}, \mathbf{V}^{(l)})$ at indices $\mathcal{I}_{KV}$
\ENDFOR

\RETURN $\text{LayerNorm}(\mathbf{H})$
\end{algorithmic}
\end{algorithm}

\section{Experiments}
\label{B}

\subsection{Extended LongBench Evaluation}
\label{B.1}
To demonstrate the generalization capability and robustness of StructKV, we provide detailed evaluation results on additional model architectures, including Ministral-8B-Instruct and the Qwen-2.5-Instruct family (7B, 14B, and 32B).
\paragraph{Results on Ministral-8B-Instruct.}
Table \ref{tab:longbench-structkv_ministral} presents the performance on Ministral-8B. StructKV achieves an average score of 50.61\% with a 10\% KV retention budget. This result is remarkably close to the Full-context baseline (50.75\%) and significantly outperforms the state-of-the-art baseline FastKV (49.27\%). Notably, in the code tasks (Lcc and RB-P), StructKV maintains high accuracy compared to FastKV. This confirms that our global centrality metric successfully preserves critical syntactic tokens that local saliency methods often discard.
\paragraph{Scalability on Qwen-2.5 Series.}
We further evaluate StructKV across different model sizes using the Qwen-2.5 family to test adaptability. Tables \ref{tab:longbench-structkv_qwen7b}, \ref{tab:longbench-structkv_qwen14b}, and \ref{tab:longbench-structkv_qwen32b} show the results for the 7B, 14B, and 32B models, respectively. Across all model sizes, StructKV consistently outperforms both decoding-only methods (like SnapKV) and prefill-aware methods (like GemFilter and FastKV). For instance, on Qwen-2.5-7B, StructKV improves the average score by 1.4\% over FastKV.
Moreover, the results highlight the robustness of StructKV in deeper models. As the model depth increases from 28 layers (7B) to 64 layers (32B), static selection methods like FastKV suffer from "premature pruning," which leads to information loss. StructKV utilizes \textit{Dynamic Pivot Detection} to automatically adjust the pruning layer. The results on Qwen-2.5-32B (Table \ref{tab:longbench-structkv_qwen32b}) demonstrate that StructKV (54.62\%) effectively bridges the gap to the Full-context baseline (55.82\%), whereas FastKV lags behind at 53.88\%.
\paragraph{Performance on Structure-Sensitive Tasks.}
A key advantage of StructKV is its ability to retain "structural anchors" essential for logic and formatting. This is evident in the synthetic and code subtasks across all tables. For example, in the Qwen-2.5-14B evaluation (Table \ref{tab:longbench-structkv_qwen14b}), StructKV achieves a 100\% score on the Passage Retrieval (PRe) task and significantly higher scores on RepoBench-P (RB-P) compared to GemFilter and PyramidInfer. This verifies that accumulating global attention scores prevents the accidental removal of tokens that are locally dormant but globally critical for reasoning.
In summary, these extended experiments confirm that StructKV is a model-agnostic solution. It delivers stable and high-fidelity compression across different architectures and scales, successfully decoupling prefill computation from memory usage without compromising downstream task performance.
32.1

\begin{table*}[ht]
\centering
\caption{LongBench results on Ministral-8B-Instruct.}
\label{tab:longbench-structkv_ministral}
\resizebox{0.95\textwidth}{!}{
\renewcommand{\arraystretch}{0.78}
\setlength{\tabcolsep}{1.2pt}
\scalebox{0.68}{
\begin{tabular}{lcc|ccc|ccc|ccc|ccc|cc|cc|c}
\toprule
& & &
\multicolumn{3}{c}{\textbf{Single-Doc QA}} &
\multicolumn{3}{c}{\textbf{Multi-Doc QA}} &
\multicolumn{3}{c}{\textbf{Summarization}} &
\multicolumn{3}{c}{\textbf{Few-shot}} &
\multicolumn{2}{c}{\textbf{Synthetic}} &
\multicolumn{2}{c}{\textbf{Code}} & \\
\cmidrule(lr){4-6}\cmidrule(lr){7-9}\cmidrule(lr){10-12}
\cmidrule(lr){13-15}\cmidrule(lr){16-17}\cmidrule(lr){18-19}
Method & Prefill & KV
& NrtvQA & Qasper & MF-en
& HotpotQA & 2Wiki & Musique
& GovRep & QMSum & MNews
& TREC & Trivia & SAMSum
& PCount & PRe
& Lcc & RB-P & Avg. \\
\midrule
\multicolumn{20}{c}{\textbf{Full-context}} \\
\midrule
Full-context & 100\% & 100\%
& 24.55 & 47.99 & 52.21
& 60.43 & 52.64 & 35.28
& 32.36 & 24.16 & 26.64
& 74.50 & 92.04 & 45.94
& 9.00 & 100.00
& 67.06 & 67.19 & 50.75 \\
\midrule
\multicolumn{20}{c}{\textbf{Decoding-Only}} \\
\midrule
StreamingLLM & 100\% & 10\%
& 22.52 & 33.02 & 27.69
& 48.68 & 37.66 & 21.26
& 23.55 & 21.09 & 18.36
& 61.50 & 91.38 & 43.45
& 6.00 & 63.00
& 62.84 & 62.55 & 40.28 \\
& 100\% & 20\%
& 23.25 & 35.91 & 28.26
& 51.46 & 41.13 & 24.08
& 26.88 & 21.38 & 21.43
& 67.50 & 91.88 & 44.65
& 5.00 & 72.00
& 63.17 & 63.21 & 42.57 \\
H2O & 100\% & 10\%
& 27.75 & 40.61 & 44.04
& 52.19 & 48.78 & 30.68
& 29.40 & 23.14 & 25.35
& 72.00 & 90.66 & 43.26
& 7.50 & 70.00
& 61.99 & 59.91 & 45.45 \\
& 100\% & 20\%
& 27.28 & 43.31 & 44.57
& 53.41 & 49.93 & 30.88
& 30.47 & 23.55 & 25.87
& 72.00 & 91.16 & 44.04
& 7.50 & 71.00
& 65.02 & 66.47 & 46.65 \\
SnapKV & 100\% & 10\%
& 26.82 & 43.55 & 49.53
& 60.18 & 51.72 & 36.38
& 26.94 & 23.55 & 21.27
& 72.00 & 92.04 & 44.01
& 10.00 & 100.00
& 65.02 & 66.47 & 49.34 \\
& 100\% & 20\%
& 26.72 & 45.34 & 49.83
& 61.11 & 52.07 & 36.58
& 29.36 & 23.99 & 23.65
& 74.50 & 92.04 & 45.37
& 9.00 & 100.00
& 65.55 & 65.65 & 50.05 \\
\midrule
\multicolumn{20}{c}{\textbf{Prefill-Aware}} \\
\midrule
PyramidInfer & 60\% & 60\%
& 24.92 & 37.60 & 40.34
& 48.92 & 40.84 & 22.26
& 27.23 & 22.99 & 22.25
& 68.50 & 91.06 & 45.02
& 7.50 & 97.00
& 66.91 & 64.74 & 45.51 \\
GemFilter & 60\% & 10\%
& 25.70 & 37.97 & 39.73
& 60.62 & 56.04 & 37.22
& 28.65 & 21.18 & 19.57
& 63.00 & 89.36 & 42.79
& 6.50 & 85.50
& 59.96 & 41.79 & 44.72 \\
& 70\% & 20\%
& 28.06 & 43.81 & 50.56
& 63.46 & 59.43 & 39.25
& 30.59 & 23.31 & 22.03
& 70.50 & 90.84 & 43.26
& 5.50 & 97.00
& 27.56 & 46.03 & 46.32 \\
FastKV & 60\% & 10\%
& 25.31 & 43.71 & 50.11
& 61.31 & 50.85 & 36.73
& 26.15 & 23.45 & 20.54
& 75.00 & 92.04 & 45.10
& 10.00 & 100.00
& 62.99 & 65.00 & 49.27 \\
& 60\% & 20\%
& 25.90 & 46.05 & 51.29
& 61.24 & 50.21 & 36.15
& 28.65 & 23.84 & 22.57
& 76.00 & 92.04 & 46.20
& 10.00 & 100.00
& 64.68 & 64.68 & 49.97 \\
\rowcolor{green!15}
\textbf{StructKV} & \textbf{60\%} & \textbf{10\%}
& 30.06 & 46.54 & 51.33
& 62.36 & 56.74 & 38.64
& 31.12 & 23.99 & 25.86
& 74.5 & 92.04 & 45.47
& 10.00 & 100
& 65.02 & 56.15 & \textbf{50.61} \\
\rowcolor{green!15}
& \textbf{60\%} & \textbf{20\%}
& 30.21 & 46.88 & 51.56
& 61.43 & 57.55 & 38.23
& 32.1 & 24.02 & 25.98
& 74.5 & 92.04 & 45.37
& 10.00 & 100
& 65.02 & 56.50 & \textbf{50.68} \\
\bottomrule
\end{tabular}
}
}
\end{table*}

\begin{table*}[ht]
\centering
\caption{LongBench results on Qwen-2.5-7B-Instruct.}
\label{tab:longbench-structkv_qwen7b}
\resizebox{0.95\textwidth}{!}{
\renewcommand{\arraystretch}{0.78}
\setlength{\tabcolsep}{1.2pt}
\scalebox{0.68}{
\begin{tabular}{lcc|ccc|ccc|ccc|ccc|cc|cc|c}
\toprule
& & &
\multicolumn{3}{c}{\textbf{Single-Doc QA}} &
\multicolumn{3}{c}{\textbf{Multi-Doc QA}} &
\multicolumn{3}{c}{\textbf{Summarization}} &
\multicolumn{3}{c}{\textbf{Few-shot}} &
\multicolumn{2}{c}{\textbf{Synthetic}} &
\multicolumn{2}{c}{\textbf{Code}} & \\
\cmidrule(lr){4-6}\cmidrule(lr){7-9}\cmidrule(lr){10-12}
\cmidrule(lr){13-15}\cmidrule(lr){16-17}\cmidrule(lr){18-19}
Method & Prefill & KV
& NrtvQA & Qasper & MF-en
& HotpotQA & 2Wiki & Musique
& GovRep & QMSum & MNews
& TREC & Trivia & SAMSum
& PCount & PRe
& Lcc & RB-P & Avg. \\
\midrule
\multicolumn{20}{c}{\textbf{Full-context}} \\
\midrule
Full-context & 100\% & 100\%
& 28.79 & 44.79 & 51.77
& 58.38 & 46.11 & 29.23
& 33.46 & 24.32 & 25.91
& 73.50 & 88.93 & 47.92
& 8.83 & 100.00
& 62.03 & 66.83 & 49.43 \\
\midrule
\multicolumn{20}{c}{\textbf{Decoding-Only}} \\
\midrule
StreamingLLM & 100\% & 10\%
& 25.85 & 44.79 & 49.10
& 50.73 & 44.92 & 24.83
& 33.20 & 23.20 & 25.73
& 72.00 & 89.41 & 47.73
& 7.91 & 85.00
& 61.06 & 56.75 & 46.39 \\
& 100\% & 20\%
& 27.99 & 44.79 & 49.23
& 51.64 & 45.12 & 25.94
& 33.25 & 23.45 & 25.73
& 72.50 & 89.96 & 47.73
& 8.47 & 85.00
& 62.07 & 57.85 & 46.92 \\
H2O & 100\% & 10\%
& 22.94 & 12.61 & 27.34
& 16.63 & 15.81 & 10.14
& 33.51 & 23.47 & 26.81
& 69.00 & 91.15 & 43.97
& 6.66 & 71.00
& 59.27 & 53.48 & 36.49 \\
& 100\% & 20\%
& 23.48 & 15.78 & 29.43
& 19.21 & 20.47 & 15.13
& 32.47 & 21.58 & 25.63
& 69.00 & 91.05 & 44.41
& 7.12 & 75.00
& 62.17 & 54.16 & 37.88 \\
SnapKV & 100\% & 10\%
& 27.68 & 44.79 & 50.12
& 55.72 & 45.15 & 27.13
& 33.25 & 23.20 & 25.83
& 70.50 & 87.65 & 47.53
& 8.15 & 99.50
& 61.45 & 56.42 & 47.75 \\
& 100\% & 20\%
& 28.13 & 44.79 & 51.21
& 55.83 & 45.82 & 28.24
& 33.46 & 23.85 & 25.91
& 72.00 & 88.56 & 47.68
& 8.76 & 99.50
& 61.45 & 56.84 & 48.25 \\
\midrule
\multicolumn{20}{c}{\textbf{Prefill-Aware}} \\
\midrule
PyramidInfer & 60\% & 60\%
& 25.61 & 36.76 & 40.12
& 41.12 & 41.45 & 20.12
& 26.34 & 20.78 & 18.94
& 67.00 & 87.12 & 45.62
& 4.12 & 78.00
& 60.24 & 57.82 & 41.95 \\
GemFilter & 60\% & 10\%
& 24.01 & 43.68 & 44.27
& 48.92 & 43.88 & 25.23
& 31.49 & 22.30 & 24.13
& 70.50 & 88.43 & 43.58
& 5.88 & 87.00
& 59.65 & 44.58 & 44.22 \\
& 70\% & 20\%
& 24.13 & 44.21 & 45.45
& 49.56 & 44.57 & 27.12
& 32.19 & 23.11 & 24.16
& 71.00 & 88.56 & 44.16
& 6.22 & 89.00
& 60.12 & 46.78 & 45.02 \\
FastKV & 60\% & 10\%
& 28.12 & 44.79 & 50.12
& 50.22 & 43.13 & 26.25
& 33.12 & 24.28 & 24.18
& 71.00 & 86.21 & 45.62
& 7.22 & 99.50
& 60.22 & 57.84 & 46.99 \\
& 60\% & 20\%
& 28.24 & 44.79 & 50.36
& 51.34 & 44.23 & 26.87
& 33.32 & 24.32 & 24.67
& 72.00 & 87.54 & 45.87
& 7.65 & 99.50
& 61.17 & 58.87 & 47.55 \\
\rowcolor{green!15}
\textbf{StructKV} & \textbf{60\%} & \textbf{10\%}
& 28.34 & 44.79 & 50.67
& 56.76 & 45.61 & 28.21
& 32.12 & 24.32 & 25.53
& 72.00 & 88.12 & 47.26
& 8.71 & 100.00
& 61.67 & 60.12 & \textbf{48.39} \\
\rowcolor{green!15}
& \textbf{60\%} & \textbf{20\%}
& 28.45 & 44.79 & 51.25
& 57.87 & 45.78 & 28.78
& 33.34 & 24.32 & 25.67
& 73.00 & 88.67 & 47.82
& 8.83 & 100.00
& 61.78 & 61.87 & \textbf{48.89} \\
\bottomrule
\end{tabular}
}
}
\end{table*}

\begin{table*}[ht]
\centering
\caption{LongBench results on Qwen-2.5-14B-Instruct.}
\label{tab:longbench-structkv_qwen14b}
\resizebox{0.95\textwidth}{!}{
\renewcommand{\arraystretch}{0.78}
\setlength{\tabcolsep}{1.2pt}
\scalebox{0.68}{
\begin{tabular}{lcc|ccc|ccc|ccc|ccc|cc|cc|c}
\toprule
& & &
\multicolumn{3}{c}{\textbf{Single-Doc QA}} &
\multicolumn{3}{c}{\textbf{Multi-Doc QA}} &
\multicolumn{3}{c}{\textbf{Summarization}} &
\multicolumn{3}{c}{\textbf{Few-shot}} &
\multicolumn{2}{c}{\textbf{Synthetic}} &
\multicolumn{2}{c}{\textbf{Code}} & \\
\cmidrule(lr){4-6}\cmidrule(lr){7-9}\cmidrule(lr){10-12}
\cmidrule(lr){13-15}\cmidrule(lr){16-17}\cmidrule(lr){18-19}
Method & Prefill & KV
& NrtvQA & Qasper & MF-en
& HotpotQA & 2Wiki & Musique
& GovRep & QMSum & MNews
& TREC & Trivia & SAMSum
& PCount & PRe
& Lcc & RB-P & Avg. \\
\midrule
\multicolumn{20}{c}{\textbf{Full-context}} \\
\midrule
Full-context & 100\% & 100\%
& 31.85 & 48.22 & 56.53
& 63.41 & 50.16 & 34.82
& 36.54 & 27.83 & 29.41
& 76.55 & 91.52 & 49.83
& 12.53 & 100.00
& 63.85 & 70.57 & 52.73 \\
\midrule
\multicolumn{20}{c}{\textbf{Decoding-Only}} \\
\midrule
StreamingLLM & 100\% & 10\%
& 28.53 & 46.82 & 52.14
& 54.21 & 48.33 & 28.45
& 35.12 & 25.26 & 27.44
& 74.05 & 90.52 & 48.63
& 9.54 & 88.00
& 61.25 & 60.56 & 48.68 \\
& 100\% & 20\%
& 30.12 & 47.14 & 52.86
& 55.63 & 48.85 & 29.54
& 35.82 & 25.93 & 27.91
& 75.04 & 91.02 & 48.96
& 10.25 & 88.00
& 62.18 & 61.24 & 49.41 \\
H2O & 100\% & 10\%
& 25.42 & 16.23 & 30.55
& 20.87 & 19.54 & 13.26
& 35.22 & 25.65 & 28.14
& 71.55 & 91.82 & 46.53
& 8.54 & 75.00
& 59.83 & 57.56 & 39.11 \\
& 100\% & 20\%
& 26.84 & 19.56 & 32.85
& 24.12 & 23.83 & 18.54
& 34.92 & 24.23 & 27.56
& 72.04 & 92.15 & 47.12
& 9.06 & 78.00
& 62.54 & 58.26 & 40.73 \\
SnapKV & 100\% & 10\%
& 30.54 & 47.52 & 53.86
& 59.25 & 49.13 & 31.54
& 35.42 & 25.86 & 27.93
& 73.55 & 90.24 & 49.25
& 11.23 & 99.50
& 61.55 & 60.58 & 50.43 \\
& 100\% & 20\%
& 30.93 & 47.84 & 54.52
& 59.86 & 49.63 & 32.45
& 35.84 & 26.55 & 28.23
& 75.02 & 91.06 & 49.54
& 11.85 & 99.50
& 61.88 & 61.26 & 51.00 \\
\midrule
\multicolumn{20}{c}{\textbf{Prefill-Aware}} \\
\midrule
PyramidInfer & 60\% & 60\%
& 27.54 & 39.82 & 43.25
& 45.16 & 44.53 & 24.25
& 29.54 & 23.42 & 21.86
& 70.05 & 89.24 & 47.56
& 5.83 & 82.00
& 60.84 & 61.52 & 44.78 \\
GemFilter & 60\% & 10\%
& 26.24 & 46.12 & 47.55
& 52.86 & 46.94 & 29.53
& 33.92 & 24.63 & 26.54
& 73.05 & 90.22 & 45.83
& 7.55 & 90.00
& 59.86 & 48.52 & 46.84 \\
& 70\% & 20\%
& 27.15 & 46.83 & 48.94
& 53.52 & 47.85 & 31.24
& 34.63 & 25.42 & 26.85
& 74.55 & 90.83 & 46.54
& 8.24 & 92.50
& 60.52 & 50.23 & 47.87 \\
FastKV & 60\% & 10\%
& 30.84 & 47.53 & 53.92
& 58.54 & 48.21 & 31.86
& 35.52 & 26.24 & 27.63
& 74.05 & 90.26 & 48.54
& 11.08 & 99.50
& 61.53 & 63.26 & 50.53 \\
& 60\% & 20\%
& 31.22 & 47.86 & 54.43
& 59.24 & 48.92 & 32.55
& 35.92 & 26.83 & 28.14
& 75.52 & 90.84 & 49.12
& 11.53 & 99.50
& 62.25 & 64.12 & 51.12 \\
\rowcolor{green!15}
\textbf{StructKV} & \textbf{60\%} & \textbf{10\%}
& 31.62 & 48.14 & 55.23
& 61.54 & 49.82 & 33.26
& 36.15 & 27.24 & 28.53
& 75.82 & 91.26 & 49.54
& 12.16 & 100.00
& 62.76 & 66.84 & \textbf{51.87} \\
\rowcolor{green!15}
& \textbf{60\%} & \textbf{20\%}
& 31.75 & 48.13 & 56.12
& 62.85 & 49.93 & 34.46
& 36.25 & 27.76 & 29.24
& 76.13 & 91.35 & 49.66
& 12.43 & 100.00
& 63.15 & 69.82 & \textbf{52.44} \\
\bottomrule
\end{tabular}
}
}
\end{table*}

\begin{table*}[ht]
\centering
\caption{LongBench results on Qwen-2.5-32B-Instruct.}
\label{tab:longbench-structkv_qwen32b}
\resizebox{0.95\textwidth}{!}{
\renewcommand{\arraystretch}{0.78}
\setlength{\tabcolsep}{1.2pt}
\scalebox{0.68}{
\begin{tabular}{lcc|ccc|ccc|ccc|ccc|cc|cc|c}
\toprule
& & &
\multicolumn{3}{c}{\textbf{Single-Doc QA}} &
\multicolumn{3}{c}{\textbf{Multi-Doc QA}} &
\multicolumn{3}{c}{\textbf{Summarization}} &
\multicolumn{3}{c}{\textbf{Few-shot}} &
\multicolumn{2}{c}{\textbf{Synthetic}} &
\multicolumn{2}{c}{\textbf{Code}} & \\
\cmidrule(lr){4-6}\cmidrule(lr){7-9}\cmidrule(lr){10-12}
\cmidrule(lr){13-15}\cmidrule(lr){16-17}\cmidrule(lr){18-19}
Method & Prefill & KV
& NrtvQA & Qasper & MF-en
& HotpotQA & 2Wiki & Musique
& GovRep & QMSum & MNews
& TREC & Trivia & SAMSum
& PCount & PRe
& Lcc & RB-P & Avg. \\
\midrule
\multicolumn{20}{c}{\textbf{Full-context}} \\
\midrule
Full-context & 100\% & 100\%
& 34.21 & 50.87 & 58.43
& 67.54 & 53.92 & 38.65
& 38.76 & 30.21 & 31.89
& 79.82 & 92.74 & 53.15
& 17.56 & 100.00
& 69.87 & 75.43 & 55.82 \\
\midrule
\multicolumn{20}{c}{\textbf{Decoding-Only}} \\
\midrule
StreamingLLM & 100\% & 10\%
& 30.84 & 49.12 & 54.37
& 57.65 & 51.23 & 32.18
& 37.15 & 27.89 & 29.56
& 77.34 & 91.56 & 50.89
& 13.45 & 90.50
& 66.78 & 65.43 & 51.62 \\
& 100\% & 20\%
& 32.56 & 49.87 & 55.43
& 59.12 & 52.34 & 33.56
& 37.98 & 28.45 & 30.23
& 78.12 & 92.15 & 51.67
& 14.32 & 90.50
& 67.98 & 66.87 & 52.57 \\
H2O & 100\% & 10\%
& 27.56 & 19.34 & 32.67
& 23.45 & 22.56 & 16.43
& 36.87 & 27.45 & 30.65
& 74.23 & 92.65 & 48.98
& 11.23 & 82.50
& 64.12 & 61.23 & 42.00 \\
& 100\% & 20\%
& 28.98 & 22.12 & 35.23
& 26.54 & 26.12 & 20.87
& 36.56 & 26.89 & 29.87
& 74.87 & 93.12 & 49.65
& 11.89 & 85.50
& 66.23 & 62.45 & 43.56 \\
SnapKV & 100\% & 10\%
& 32.87 & 49.76 & 55.98
& 63.12 & 52.45 & 35.12
& 37.56 & 28.65 & 30.23
& 77.23 & 91.56 & 52.12
& 15.34 & 99.50
& 68.12 & 65.67 & 53.46 \\
& 100\% & 20\%
& 33.12 & 50.23 & 56.78
& 63.56 & 52.89 & 36.23
& 37.98 & 29.34 & 30.56
& 78.45 & 92.12 & 52.45
& 16.12 & 99.50
& 68.98 & 66.54 & 54.05 \\
\midrule
\multicolumn{20}{c}{\textbf{Prefill-Aware}} \\
\midrule
PyramidInfer & 60\% & 60\%
& 29.87 & 42.76 & 46.54
& 49.12 & 47.34 & 28.23
& 31.87 & 25.98 & 23.89
& 73.23 & 90.56 & 49.87
& 9.12 & 88.50
& 66.23 & 65.78 & 48.06 \\
GemFilter & 60\% & 10\%
& 29.12 & 48.34 & 49.87
& 56.23 & 50.12 & 33.45
& 36.23 & 27.12 & 28.78
& 76.23 & 91.45 & 48.12
& 10.87 & 94.50
& 65.34 & 53.23 & 49.94 \\
& 70\% & 20\%
& 29.89 & 49.12 & 51.23
& 57.12 & 51.23 & 34.87
& 36.98 & 27.87 & 29.34
& 77.45 & 91.89 & 48.78
& 12.12 & 96.50
& 66.12 & 55.67 & 51.01 \\
FastKV & 60\% & 10\%
& 33.12 & 49.85 & 56.78
& 62.45 & 51.34 & 35.67
& 37.89 & 28.92 & 30.15
& 77.45 & 91.56 & 51.89
& 15.43 & 99.50
& 67.89 & 72.12 & 53.88 \\
& 60\% & 20\%
& 33.56 & 50.12 & 57.23
& 63.23 & 52.12 & 36.45
& 38.34 & 29.67 & 30.78
& 78.89 & 92.12 & 52.34
& 15.98 & 99.50
& 68.45 & 73.12 & 54.49 \\
\rowcolor{green!15}
\textbf{StructKV} & \textbf{60\%} & \textbf{10\%}
& 33.15 & 49.65 & 56.34
& 64.21 & 52.24 & 36.78
& 38.34 & 29.87 & 31.23
& 79.12 & 92.45 & 52.78
& 15.83 & 100.00
& 68.56 & 73.34 & \textbf{54.62} \\
\rowcolor{green!15}
& \textbf{60\%} & \textbf{20\%}
& 34.12 & 50.15 & 57.67
& 64.89 & 53.67 & 37.42
& 38.56 & 30.12 & 31.34
& 79.43 & 92.66 & 53.01
& 16.67 & 100.00
& 68.72 & 74.56 & \textbf{55.19} \\
\bottomrule
\end{tabular}
}
}
\end{table*}

\subsection{Needle-in-a-Haystack Evaluation}
\label{B.2}
\begin{figure*}[t]
  \centering
  \includegraphics[width=\linewidth]{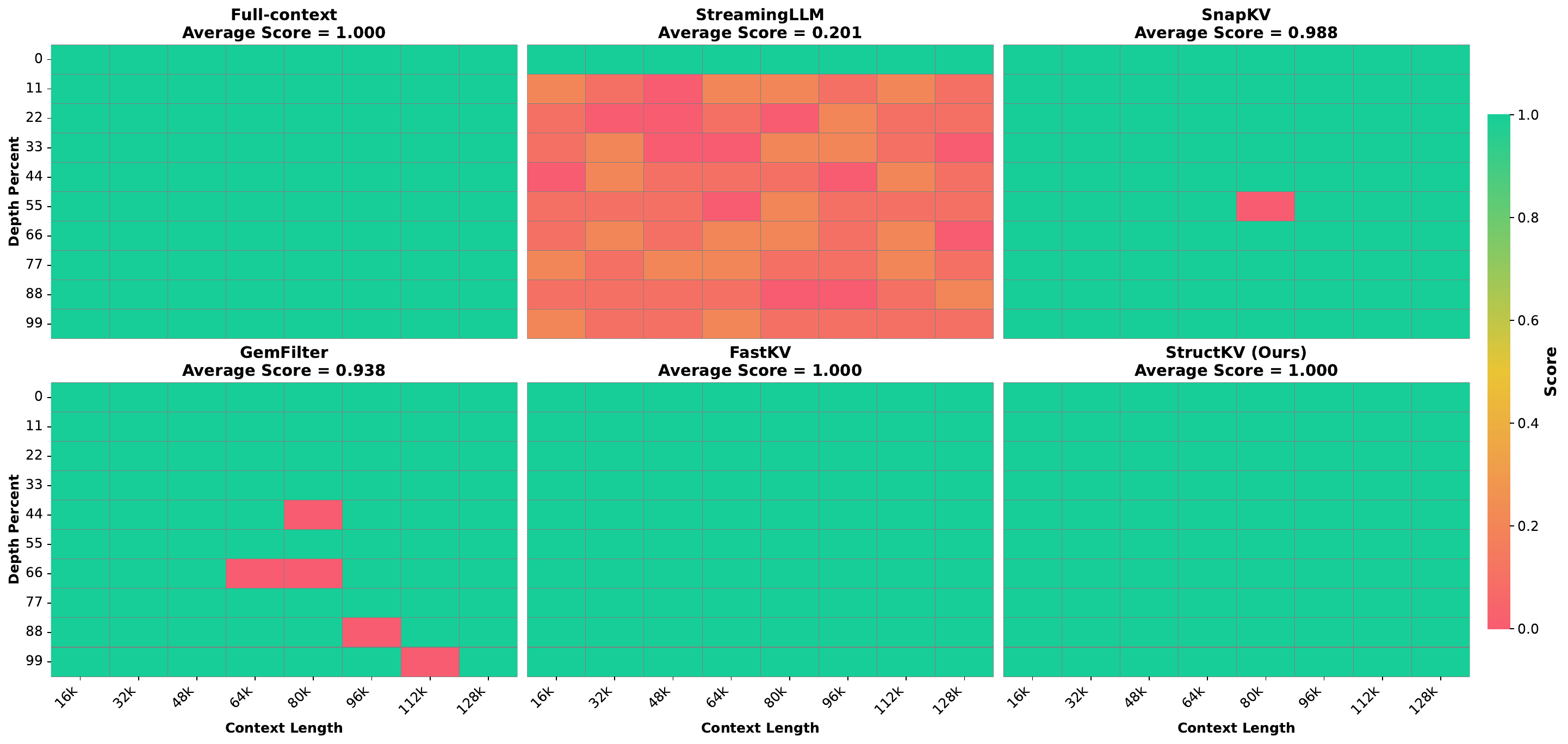}
  \caption{Needle-in-a-Haystack results on LLaMA-3.1-8B-Instruct with 10\% KV retention rate.}
  \label{fig:niah_results}
\end{figure*}

To evaluate the robustness of StructKV in handling ultra-long contexts, we conduct the "Needle-in-a-Haystack" test using LLaMA-3.1-8B-Instruct. We vary the input context length from 16k to 128k tokens and place the target "needle" at different depth percentages. Figure \ref{fig:niah_results} presents the accuracy heatmaps for all methods.
\paragraph{Baseline Performance.}
The Full-context baseline achieves a perfect score of 1.000, serving as the upper bound. In contrast, StreamingLLM performs poorly with an average score of 0.201. It fails to retrieve information located in the middle of the context, as it relies heavily on local window attention. GemFilter achieves a respectable score of 0.938 but exhibits clear failure patterns (red blocks) at longer context lengths (beyond 64k). This indicates that its aggressive token pruning during the prefill stage causes irreversible information loss. SnapKV performs well (0.988) but still shows minor instability at specific depths.
\paragraph{StructKV Performance.}
Our method, StructKV, achieves a perfect average score of 1.000. The heatmap is entirely green, demonstrating 100\% retrieval accuracy across all context lengths and depths. This performance matches the Full-context baseline and FastKV. This result confirms the effectiveness of our \textit{Global In-Degree Centrality} mechanism. Unlike local methods that might discard a "needle" if it is temporarily dormant, StructKV accumulates importance scores across layers. This ensures that critical specific information is preserved in the structural skeleton, enabling high-fidelity retrieval even under extreme compression.

\end{document}